\documentclass{article}

\usepackage{arxiv}

\usepackage{ulem}
\usepackage[utf8]{inputenc} 
\usepackage[T1]{fontenc}    
\usepackage{hyperref}       
\usepackage{url}            
\usepackage{booktabs}       
\usepackage{amsmath, amsfonts}       
\usepackage{nicefrac}       
\usepackage{microtype}      
\usepackage{lipsum}		
\usepackage{graphicx}
\usepackage[numbers,sort&compress,square]{natbib}
\usepackage{doi}

\usepackage{multirow}
\usepackage{lineno}
\usepackage{soul}
\usepackage{comment}

\graphicspath{{./Figures-png/}}

\usepackage{xcolor}

\usepackage{subcaption}

\title{From classification to regression: using a fruitfly to solve equations}

\author{ 
    \href{https://orcid.org/0000-0001-5548-0265}{\includegraphics[scale=0.06]{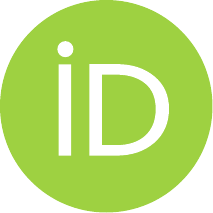}
    \hspace{1mm}Shady E. Ahmed} \\
	Advanced Computing, Mathematics, and Data Division \\
	Pacific Northwest National Laboratory\\
	Richland, WA 99354 \\
	\texttt{shady.ahmed@pnnl.gov} \\
    \And
    \href{https://orcid.org/0000-0002-9928-5637}{\includegraphics[scale=0.06]{orcid.pdf}
    \hspace{1mm}Panos Stinis} \\
	Advanced Computing, Mathematics, and Data Division \\
	Pacific Northwest National Laboratory\\
	Richland, WA 99354 \\
  	\texttt{panagiotis.stinis@pnnl.gov} \\
}

\renewcommand{\shorttitle}{Using a Fruitfly to Solve Equations}

\hypersetup{
pdftitle={A template for the arxiv style},
pdfsubject={q-bio.NC, q-bio.QM},
pdfauthor={David S.~Hippocampus, Elias D.~Striatum},
pdfkeywords={},
}

\begin{document}
\maketitle

\begin{abstract}
	We present a novel approach to regression tasks using classification which is motivated by the mechanism used by  fruitflies to sense their environment. Specifically, we formulate a general framework for learning nonlinear input-output relationships by replacing complex global surrogate models with a finite library of representative local patterns. Since scientific data often occupy limited and recurring regions of the input space, we generate predictions by measuring similarities between a query and stored patterns, then combining their associated responses through weighted reconstruction. We apply this approach to nonlinear dynamical systems, data-driven regression, and physics-informed learning using suitable embeddings and similarity measures. For dynamical systems, our offline-online workflow extracts patterns from data or governing equations during the offline phase, while online prediction requires only similarity evaluation and response aggregation. This structure helps us reduce computational and memory demands while providing explicit control over the trade-off among accuracy, storage, and inference cost.
\end{abstract}


\section{Introduction} \label{sec:intro}
Learning nonlinear input-output relationships is a fundamental problem across scientific computing. Examples include predicting the evolution of nonlinear dynamical systems, constructing surrogate models for expensive simulations, and solving regression problems from experimental or simulation data. Despite their diverse applications, these problems share a common objective: given an input representation, infer the corresponding response through an unknown, potentially nonlinear, mapping.

A prevailing strategy is to approximate this mapping directly using a global surrogate, e.g., reduced-order models \cite{lucia2004reduced,benner2017model,rowley2017model}, kernel methods \cite{hofmann2008kernel,shawe2004kernel}, Gaussian processes \cite{williams1995gaussian,schulz2018tutorial,wang2023intuitive}, radial basis function interpolation \cite{buhmann2000radial,flyer2009radial}, and deep neural networks \cite{bishop2023deep, prince2023understanding,chollet2021deep}. The success of these methods has enabled significant advances in scientific machine learning and surrogate modeling. However, these methods typically learn a globally parameterized mapping whose expressive power is determined by the capacity of the underlying network architecture. As the underlying relationship becomes increasingly complex, accurate global approximations typically require larger models, higher inference costs, and greater memory requirements. This severely limits the suitability of these approaches for energy-constrained applications, such as edge computing, embedded sensing, and autonomous platforms. In these settings, inference cost is often as important as predictive accuracy, motivating surrogate models that can achieve competitive performance while requiring substantially fewer floating-point operations and memory accesses.

An important observation, however, is that many scientific datasets do not uniformly populate their ambient input space \cite{fefferman2013testing}. For example, trajectories of nonlinear dynamical systems generally explore only a small portion of the admissible state space and often revisit recurring dynamical regimes \cite{packard1980geometry}. Similarly, many regression problems exhibit locally coherent behavior that can be well represented through a finite collection of representative patterns. These observations suggest that accurate prediction may not require learning a globally valid approximation of the underlying mapping. Instead, it may be sufficient to identify and combine a small number of representative local behaviors. This principle underlies a broad class of techniques, including local regression \cite{cleveland1988locally}, moving least-squares methods \cite{levin1998approximation,mirzaei2015analysis}, partition-of-unity approaches \cite{melenk1996partition,babuvska1997partition,fan2023probabilistic}, and piecewise-affine models \cite{ferrari2003clustering,breschi2016piecewise,bemporad2022piecewise}, where the input space is decomposed into regions and separate approximations are constructed within each region \cite{amsallem2012nonlinear,ahmed2020breaking}. Such methods can significantly improve approximation accuracy by exploiting local regularity, but they typically require solving or evaluating multiple local regression models during inference, and their complexity often increases with the number of local regions or training samples.

Motivated by these observations and inspired by biological systems which routinely perform complex inference using remarkably limited computational resources e.g., the mechanism through which a  fruitfly senses its environment \cite{liang2021can}, we propose a different representation of regression. Rather than approximating the entire input-output mapping with a single global model, we represent it through a finite collection of representative patterns. Given a query, similarities between it (or its embedding) and the stored patterns are used to activate the most relevant local behaviors, and the desired output is reconstructed as a weighted combination of their associated responses.

The framework is intentionally general and the proposed formulation is independent of the underlying regression task. For nonlinear dynamical systems, the similarity is evaluated between the current system state (or features derived from it) and the representative patterns. For conventional regression, an embedding may instead be constructed from the independent variables using prescribed basis functions, learned feature maps, or other problem-dependent representations \cite{sauer1991embedology}. This perspective unifies dynamical prediction, supervised regression, and physics-informed learning within a common framework.

This approach naturally admits an offline-online decomposition. During offline training, representative patterns and their associated outputs are extracted from available data or governing equations. Online prediction requires only computing similarities in the embedding space, and aggregating a small number of stored responses. Consequently, the online computational cost depends primarily on the size and structure of the learned pattern library rather than on evaluating a complex global surrogate. Moreover, the framework provides explicit control over the trade-off between accuracy, memory consumption, and inference cost through the number of stored patterns, the embedding, and the similarity metric.

The contributions of this work are as follows:
\begin{itemize}
    \item We introduce a general pattern-learning and activation framework that reformulates regression as weighted reconstruction over representative patterns.
    \item For dynamical systems, we show that the learned patterns can efficiently reconstruct continuous- or discrete-time evolution operators.
    \item We extend the same formulation to conventional regression problems using embeddings constructed from the independent variables, enabling both data-driven and physics-informed formulations within a unified framework.
\end{itemize}

As discussed above, the idea that complex functions can be easier to represent locally than globally has been a cornerstone for local regression techniques. However, the perspective adopted in this work is fundamentally different. Rather than partitioning the state space to fit multiple local models, we seek to identify a small collection of representative dynamical patterns that capture recurring system behaviors. Prediction is then performed by classifying the current state according to its similarity to these learned patterns and combining the associated representative responses. Consequently, the computational burden shifts from repeated local regression toward similarity-based pattern selection, yielding a compact representation whose complexity depends primarily on the number of learned patterns rather than the size of the training data. The proposed framework is also distinct from mixture-of-experts architectures \cite{jacobs1991adaptive} and related adaptive regression models where a gating mechanism selects or combines multiple expert predictors \cite{shazeer2017outrageously}, each of which represents an independently parameterized regression model \cite{sharma2024ensemble}. In mixture-of-experts frameworks, prediction requires evaluating one or more experts after the routing decision has been made. In contrast, our approach does not rely on multiple regression models. Instead, similarity to representative patterns directly determines how previously learned representative responses are combined.

The remainder of this paper is organized as follows. Section~\ref{sec:formulation} presents the problem formulation for both dynamical systems prediction and function regression. Section~\ref{sec:method} introduces the proposed pattern-learning framework and describes its different variants. Numerical results for dynamical systems are presented in Section~\ref{sec:res-dynamics}, where the framework is demonstrated on the Lotka-Volterra and Lorenz systems. Section~\ref{sec:res-regression} evaluates the proposed approach on classical regression problems, including both data-driven and physics-informed learning settings. Finally, Section~\ref{sec:conc} concludes the paper with a discussion of the main findings, limitations, and future research directions.

\section{Problem Formulation} \label{sec:formulation}
We begin by considering nonlinear dynamical systems, which provide the primary motivation for the proposed framework. We then show that the resulting regression formulation naturally extends to more general function approximation problems.

\subsection{Nonlinear Dynamical Systems}
Consider an autonomous dynamical system governed by
\begin{equation}
    \frac{du}{dt}=f(u),
\end{equation}
or, equivalently, by the discrete-time evolution map
\begin{equation}
    u(t+\Delta t)=\mathcal{M}(u(t)),
\end{equation}
where $u\in\mathbb{R}^{n}$ denotes the system state.

Given the current state, the objective is to predict either the instantaneous time derivative $f(u)$, or the flow map $\mathcal{M}(u)$, depending on the chosen formulation. 

Most existing surrogate models seek to approximate $f(\cdot)$ or $\mathcal{M}(\cdot)$ using a single global function. While successful in many applications, this approach may become challenging when the underlying dynamics exhibit strong nonlinearities across the admissible state space. We note that the trajectories of many nonlinear dynamical systems do not uniformly explore the state space. Instead, they typically remain confined to low-dimensional subsets and repeatedly revisit regions characterized by similar dynamical behavior. Consequently, nearby states often produce similar responses, suggesting that prediction may be achieved by identifying and combining representative local behaviors rather than relying on a globally valid approximation. This observation motivates the representation proposed in this work, as described later in Section~\ref{sec:method}.

 \subsection{Extension to General Regression}\label{sec:general regression}
Although the proposed framework is motivated by nonlinear dynamical systems, the underlying formulation is not restricted to state-space models. Consider the general regression problem
\begin{equation}
    u = F(x),
\end{equation}
where $x$ denotes one or more independent variables. Unlike dynamical systems, the inputs need not correspond to physical states or possess an inherent temporal structure. Instead, we introduce an embedding $\phi$ which maps the original inputs into a feature space,
\begin{equation}
    y=\phi(x).
\end{equation}
The embedding may be constructed from prescribed basis functions, finite-element shape functions, Fourier features, radial basis functions, learned latent representations, or other problem-dependent transformations. The proposed framework does not assume any particular choice of embedding, provided that it captures the local structure of the regression problem. Once the embedding has been constructed, regression is performed entirely in the embedding space. As will be shown in the following section, both nonlinear dynamical systems and conventional regression can therefore be treated within a common computational framework, differing only in the construction of the embedding.

\section{Method} \label{sec:method}




The central idea of this work is to reformulate regression as a classification problem. At first glance, these two learning tasks appear fundamentally different. Classification partitions an input space into a finite number of categories, whereas regression predicts continuous-valued outputs. We argue, however, that this distinction stems primarily from how the output of the classifier is interpreted \cite{torgo1996regression,salman2012regression}. The proposed framework reformulates regression as a classification problem by interpreting class activations as input-dependent reconstruction coefficients rather than class probabilities. Instead of learning a single global approximation of the target mapping, the model learns a collection of representative local patterns together with their associated responses. Prediction then proceeds by associating the current input to these representative patterns and reconstructing the desired output through their weighted combination.

We first present the formulation for nonlinear dynamical systems, where the notion of recurring dynamical patterns provides the primary motivation for the proposed approach. We then show that the same framework naturally extends to general regression problems through an appropriate choice of input embedding.

\subsection{Learning Continuous-Time Dynamics}
Consider the nonlinear dynamical system
\begin{equation}
    \dfrac{\mathrm{d}u}{\mathrm{d}t} = f(u), 
\end{equation}
where $u(t)\in\mathbb{R}^{n}$ denotes the system state and $f:\mathbb{R}^{n}\rightarrow\mathbb{R}^{n}$ is the governing right-hand-side operator. Rather than approximating the governing operator through a single global nonlinear model, we seek a compact collection of representative dynamical patterns that characterize these recurring local behaviors. Therefore, we introduce a pattern library
\begin{equation}
\mathcal{P} = \{v_k\}_{k=1}^{P},
\end{equation}
where $v_k\in\mathbb{R}^{n}$ denotes the representative $k^{th}$ pattern and $f(v_k)$ denotes the corresponding dynamical response. Although the idea of representing complex datasets using a limited number of prototypes has been extensively explored in clustering \cite{kaiser2013cluster,fernex2021cluster} and related representation learning methods, the learned patterns in the present work serve a fundamentally different purpose. They are not introduced to reconstruct the observed data or discover latent structure, but to represent recurring modes of system behavior that directly support prediction. 

The learned patterns therefore become computational primitives that connect classification with regression, rather than merely compressed summaries of the training set. Given a query state $u(t)$, its similarity to every representative pattern is computed using an inner product as follows:
\begin{equation}
s_k(u(t))= \langle u(t) , v_k \rangle.
\end{equation}
The similarities are transformed into normalized activations using the softmax function,
\begin{equation}
w_k(u(t)) = \frac{\exp(\tau s_k)}
{\displaystyle\sum_{j=1}^{P}\exp(\tau s_j)},
\end{equation}
where $\tau>0$ is a temperature parameter, which can be learnable or fixed, e.g., $\dfrac{1}{\sqrt{n}}$. The approximate right-hand-side operator is then reconstructed as
\begin{equation}
\hat{f}(u(t)) = \sum_{k=1}^{P} w_k(u(t)) f(v_k), \label{eq:rhs_reconstruction}
\end{equation}
where $f(v_k)$ can be pre-computed and stored during the offline stage. Equation~(\ref{eq:rhs_reconstruction}) constitutes the central idea of the proposed framework. Instead of directly evaluating a globally defined nonlinear surrogate, prediction consists of classifying the current state among a finite collection of representative dynamical patterns and reconstructing the governing operator from their associated responses. Since the weights $w_k$ are nonnegative and satisfy $\sum_{k=1}^{P} w_k=1$, the predicted response is a convex combination of the representative responses. Consequently, predictions are confined to the convex hull of the learned responses, providing an inherently bounded approximation despite limiting extrapolation beyond the behaviors represented by the learned patterns.

Our approach also naturally admits a top-$R$ sparse formulation by retaining only the $R$ most significant patterns and setting the remaining weights to zero during inference. In particular, one can define $T_R$ as the index set of top $R$ similarity scores, $T_R = \{s_k | s_k>\epsilon, k=1, \ldots, P \}$ for a threshold $\epsilon$. The associated weights are re-normalized as follows:
\begin{equation}
w_k(u(t)) = \begin{cases}
    \dfrac{\exp(\tau s_k)}
{\displaystyle\sum_{j \in T_R}\exp(\tau s_j)}, \qquad  &k\in T_k,\\
0, \qquad &otherwise.
\end{cases}
\end{equation}

The reconstructed operator replaces the original right-hand side within any standard numerical time integration scheme allowing trajectories to be advanced using explicit or implicit integrators in exactly the same manner as the original governing equations. Two supervision settings can be considered:
\paragraph{Direct operator supervision.}
When evaluations of the governing operator are available, we can evaluate $f(v_k)$ for any given pattern $v_k$. Thus, the pattern library can be learned from the dataset
\begin{equation}
\mathcal{D}
=
\left\{
(u_i,f(u_i))
\right\}_{i=1}^{N},
\end{equation}
by minimizing
\begin{equation}
\mathcal{L}_{\mathrm{RHS}}
=
\frac{1}{N}
\sum_{i=1}^{N}
\left\|
f(u_i)-\hat{f}(u_i)
\right\|_2^2.
\end{equation}
This setting directly supervises the approximation of the governing operator, and the system's dynamics can be approximated using Eq.~\ref{eq:rhs_reconstruction}. Alternatively, Eq~\ref{eq:rhs_reconstruction} could be fed into a time integration scheme to produce roll-outs of arbitrary lengths which are then compared against ground truth trajectories using suitable loss metrics (e.g., mean-squared error), akin to in-the-loop training \cite{ahmed2023multifidelity}.

\paragraph{Trajectory supervision.}
In many practical applications, evaluations of $f$ are unavailable and only sampled trajectories are observed. In this case, a surrogate model (e.g., neural network with trainable parameters $\theta$) is used to approximate $f$ as follows:
\begin{equation}
\hat{f}(u(t)) = \sum_{k=1}^{P} w_k(u(t)) f^{\theta}(v_k), \label{eq:rhs_reconstruction_unknwonf}
\end{equation}
and the reconstructed operator is integrated forward in time and the resulting trajectories are compared against the observations through
\begin{equation}
\mathcal{L}_{\mathrm{traj}} = \sum_{n} \left\| u(t_n)-\hat{u}(t_n) \right\|_2^2.
\end{equation}
This formulation resembles Neural ODEs in its optimization objective while employing a fundamentally different representation of the underlying dynamics. Once trained, only the patterns $v_p$ and their associated approximate responses $f^{\theta}(v_k)$ are stored to minimize the memory footprint, i.e., no need to store/load the weights $\theta$ of the neural network.

\subsection{Learning Discrete-Time Flow Map}
The proposed framework extends naturally to discrete-time dynamical systems,
\begin{equation}
u(t+\Delta t) = \mathcal{M}(u (t)),
\end{equation}
where $\mathcal{M}$ denotes the one-step flow map. Rather than storing representative evaluations of the continuous-time right-hand side, the pattern library now stores representative one-step state transitions, $\mathcal{M}(v_k)$. After computing the pattern activations, the one-step evolution is reconstructed as
\begin{equation}
\hat{\mathcal{M}}(u(t)) = \sum_{k=1}^{P} w_k (u(t)) \mathcal{M}(v_k).
\end{equation}
We note that $\mathcal{M}(v_k)$ is evaluated only during the offline phase either using the exact flow map (when system's dynamics is known) or through an emulator $\mathcal{M}^{\theta}(v_k)$, then stored to be used in the online phase. Long-term predictions are obtained through recursive application of the learned flow map,
\begin{equation}
u(t+\Delta t) = \hat{\mathcal{M}}(u(t)).
\end{equation}

Compared with the continuous-time formulation, this approach eliminates the need for numerical time integration during inference but is specific to the time step used during training.

\subsection{Extension to General Regression} \label{sec:function_regression}

Although motivated by nonlinear dynamical systems, the proposed framework is not restricted to state-space models. Consider the general regression problem
\begin{equation}
u = F(x),
\end{equation}
where $x$ denotes one or more independent variables. Unlike dynamical systems, no natural notion of state or dynamical pattern
exists. We therefore introduce an embedding
\begin{equation}\label{embedding}
y = \phi(x),
\end{equation}
where $\phi$ maps the independent variables into a feature space in which local patterns can be represented. 
While in future work we will explore more general representations, here we will use a representation which is directly motivated by the fruitfly's olfactory system \cite{liang2021can}. Specifically, we assume that the query point $x_Q$ belongs in a domain $D.$ We populate the domain $D$ with $P$ points whose locations are chosen randomly, uniformly or can even be trainable. The values of the unknown quantity $u$ at these $P$ points play the same role as the patterns in the preceding discussion about dynamical systems. For each of the $P$ (pattern) points, we associate a normally distributed influence function (centered at the point and with a user prescribed variance), which can be thought of as the point's "scent". Similarly, for each query point, we can consider an  influence function (centered at the point and with a user prescribed variance), which will be the query point's scent. For each query point and each of the pattern points, we can sample their influence functions at a prescribed set of points within the domain $D,$ say $n,$ and form $n$-dimensional vectors. Then, we can use these $n$-dimensional vectors with a softmax function as in the previous discussion to find the similarity between the query point and each of the pattern points. The softmax weights are then used to prescribe the prediction for the value of $u$ at the query point $x_Q$ as a linear combination of the $P$ pattern values.  

To be concrete, let us assume that the domain $D$ is one-dimensional and we represent the scent of a query point $x_Q$ as
\begin{equation}\label{query_scent}
\phi_Q(x)=\frac{1}{\sqrt{2\pi}\sigma_Q}\exp{[-\frac{(x-x_Q)^2}{2\sigma_Q^2}]}.
\end{equation}
Similarly, we represent the scent of the $k$-th pattern point $x_{Pk},$ where $k=1,\ldots,P,$ as
\begin{equation}\label{pattern_scent}
\phi_{Pk}(x)=\frac{1}{\sqrt{2\pi}\sigma_P}\exp{[-\frac{(x-x_{Pk})^2}{2\sigma_P^2}]}.
\end{equation}
For the query and pattern points, we can form the $n$-dimensional vectors
\begin{equation}\label{query_lift}
v_Q=[\phi_Q(z_1),\ldots,\phi_Q(z_n)]^T
\end{equation}
and 
\begin{equation}\label{pattern_lift}
v_{Pk}=[\phi_{Pk}(z_1),\ldots,\phi_{Pk}(z_n)]^T,
\end{equation}
where $z_1,\ldots,z_n$ are points in $D.$
Then, we can compute the softmax weights $w_k(x_Q),$ for $k=1,\ldots,P$ as
\begin{equation}\label{softmax_regression}
w_k(x_Q)=\frac{\exp{(v_Q^T v_{Pk})}}{\sum_{k=1}^P \exp{(v_Q^T v_{Pk})}}
\end{equation}
In addition, let us associate with the $k$-th pattern point $x_{Pk}$, the trainable pattern value $\tilde{u}_{Pk}.$ Then, the prediction of our fruitfly regressor for the query point $x_Q$ is given by
\begin{equation}\label{regressor_prediction}
\hat{u}(x_Q)=\sum_{k=1}^P w_k(x_Q) \tilde{u}_{Pk}.
\end{equation}
The fruitfly regressor \eqref{regressor_prediction} estimates the function value at $x_Q$ as a weighted combination of the trainable patterns $\tilde{u}_{Pk}.$

If we have $N$ training pairs $\{x_i,u_i\}_{i=1}^N,$ then we can use them to estimate the trainable pattern values $\tilde{u}_{Pk}$ by minimizing the loss function
\begin{equation}
\label{regressor_loss}
\mathcal{L}=\frac{1}{N} \sum_{i=1}^N [u_i -\hat{u}(x_i) ]^2.
\end{equation}

If instead of data training pairs, we only have access to the equation satisfied by the unknown function $u(x),$ we can proceed with the same representation as in \eqref{regressor_prediction}, but the loss function is replaced by a physics-informed one, as it is done in physics-informed neural networks \cite{raissi2019physics}.


Although the proposed formulation shares superficial similarities with kernel methods and Nadaraya-Watson estimators \cite{cai2001weighted} which produce predictions through weighted combinations of reference quantities \cite{larsson2017least}, the role of these weights is fundamentally different. In classical kernel methods, weights are computed with respect to individual training samples or predefined basis centers, and the approximation typically depends directly on the entire training dataset. In contrast, our method first learns a compact set of representative patterns that summarize the dominant behaviors observed during training. Similarity is therefore evaluated with respect to these learned patterns rather than the original data, allowing the predictive model to remain compact even as additional training samples become available. Nonetheless, our approach can benefit from theoretical guarantees and theorems that have been developed in these areas, a topic to be explored in future studies.

\section{Applications to Dynamical Systems} \label{sec:res-dynamics}
We first illustrate the proposed framework using two prototypical dynamical systems to demonstrate its capabilities in representative settings. In the first example, we consider the Lotka-Volterra system to examine scenarios in which (1) the governing equations are known and (2) the governing equations are unavailable. For the latter case, the proposed framework can learn either the system dynamics by approximating the right-hand side of the governing equations or the corresponding one-step flow map directly from data. In the second example, we consider the Lorenz system to demonstrate how the framework identifies representative local patterns to enable accurate long-term prediction of chaotic dynamics. We further show how a continual learning strategy can extend the prediction horizon by incrementally incorporating newly observed patterns as additional data become available.

\subsection{Lotka-Volterra System}
The Lotka-Volterra system is a classical nonlinear dynamical model describing the interaction between prey and predator populations. It serves as a canonical benchmark for nonlinear systems due to its low dimensionality, nonlinear coupling, and oscillatory dynamics. The governing equations are
\begin{equation}
\begin{aligned}
\frac{dx}{dt} &= \alpha x - \beta xy,\\
\frac{dy}{dt} &= \delta xy - \gamma y,
\end{aligned} \label{eq:lotka-volterra}
\end{equation}
where $x(t)$ and $y(t)$ denote the prey and predator populations, respectively, and $\alpha$, $\beta$, $\delta$, and $\gamma$ are positive parameters governing prey growth/birth rate, predation interaction strength, predator growth rate, and predator death rate. In this work, we use values of $\alpha=1.1$, $\beta=0.4$, $\delta=0.1$, and $\gamma=0.4$, , which yield nonlinear oscillatory dynamics with a fixed point at $(x,y)_{\text{fixed}}=(4,2.75)$. Solution trajectory is generated from an initial condition of $(x,y)_{\text{IC}}=(10,5)$ through numerical integration of the governing equations.

\subsubsection{Learning with known system dynamics}
We first consider the case in which the governing equations of the Lotka-Volterra system are known; that is, the right-hand side of Eq.~\ref{eq:lotka-volterra} can be evaluated at any pattern $v_k\in\mathbb{R}^2$. This represents the simplest setting of the proposed pattern identification framework. We apply the methodology described in Section~\ref{sec:method} using varying numbers of patterns to investigate the trade-off between model complexity, prediction accuracy, and the representativeness of the identified patterns.

\begin{figure}[ht!]
    \centering
    \includegraphics[width=\linewidth]{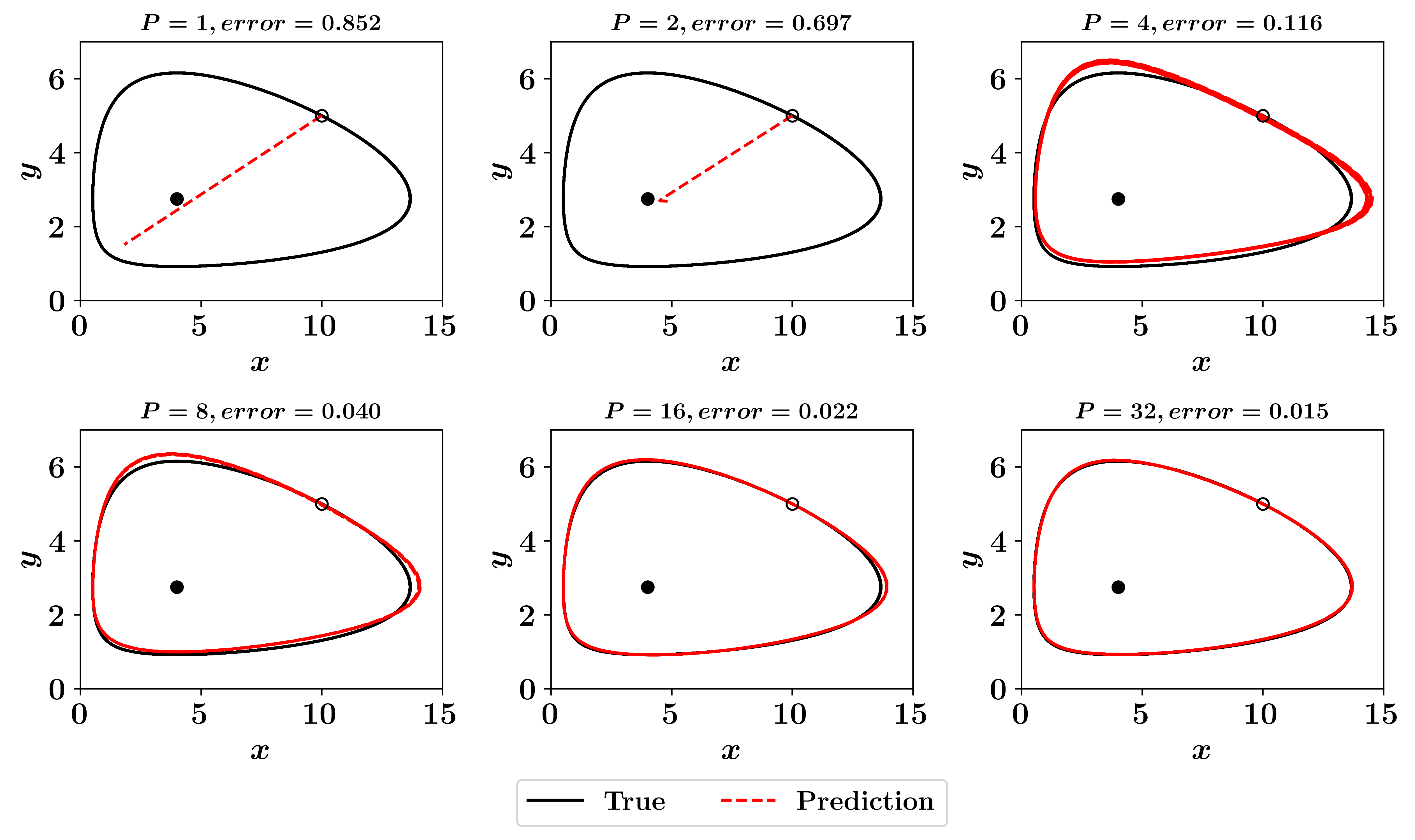}
    \caption{Prediction of the Lotka-Voletrra dynamics using different numbers of patterns along with the associated relative $\ell_2$ error. The black circle/dot indicates the system's fixed point.}
    \label{fig:lotka-known-rhs-phase}
\end{figure}

Figure~\ref{fig:lotka-known-rhs-phase} presents the resulting phase portraits for different numbers of patterns. As expected, increasing the number of patterns yields a progressively more accurate approximation of the underlying dynamics, with the approximation error decreasing monotonically as the model complexity increases. The resulting model remains highly compact: it stores only $nP$ parameters, where $n$ is the state dimension ($n=2$ for the Lotka-Volterra system) and $P$ is the number of patterns. This parameter count is substantially smaller than that of conventional global approximation methods, such as neural networks, which typically require significantly more parameters to represent the dynamics over the entire state space.

\begin{figure}[ht!]
    \centering
    \includegraphics[width=\linewidth]{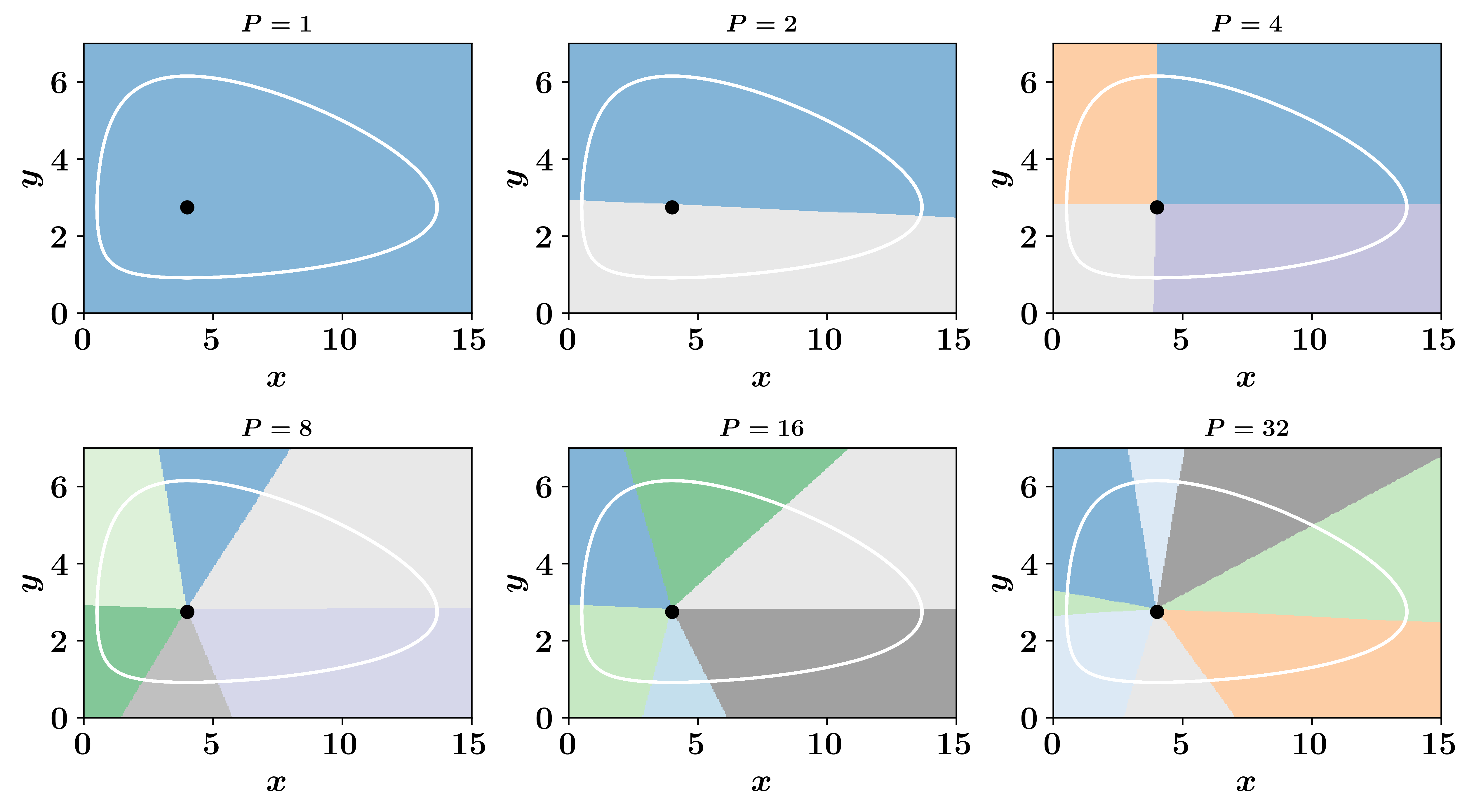}
    \caption{Pattern dominance regions for the Lotka-Volterra system. Each point in the state space is colored according to the index of the pattern with the largest similarity score. The black circle/dot indicates the system's fixed point.}
    \label{fig:lotka-known-rhs-ownership}
\end{figure}

\begin{figure}[ht!]
    \centering
    \includegraphics[width=\linewidth]{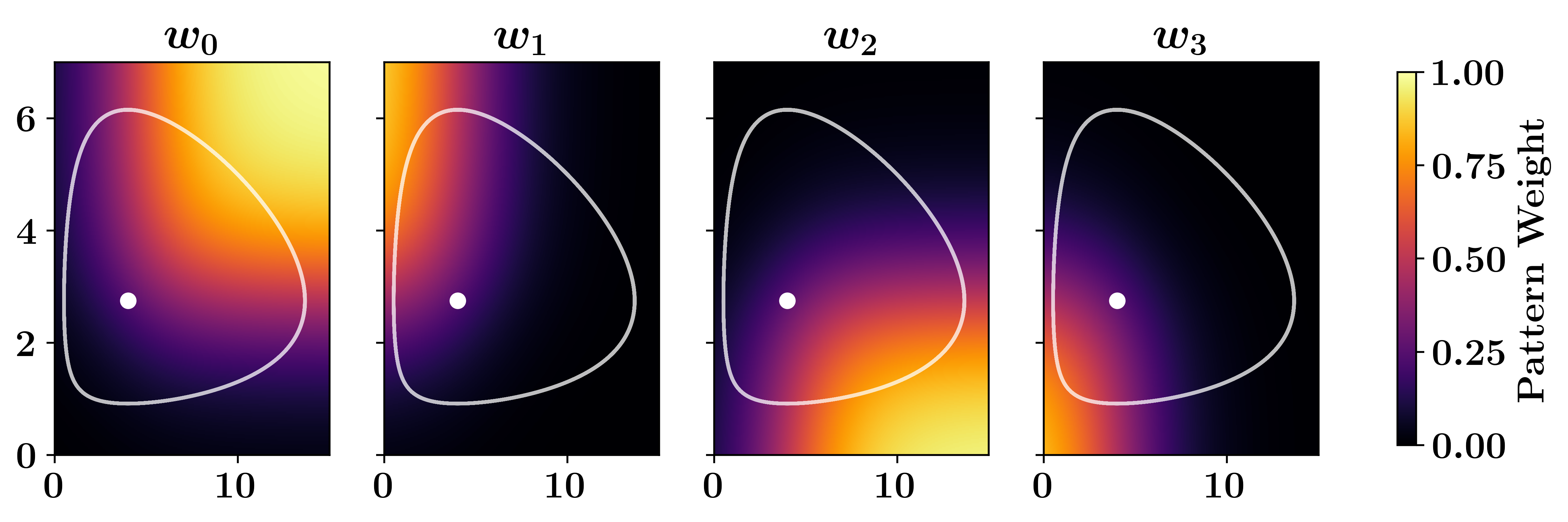}
    \caption{Assigned weight for each pattern in different regions of the state space for $P=4$.}
    \label{fig:lotka-known-rhs-pattern-weight}
\end{figure}

Figure~\ref{fig:lotka-known-rhs-ownership} visualizes the regions of the state space associated with each identified pattern. Specifically, each point is colored according to the index of the pattern with the largest similarity score,
$\arg\max_k \left(v_k^\top u\right)$, thereby partitioning the state space into regions of pattern dominance. Several observations can be made. First, the identified patterns induce a clear partition of the state space, with each region corresponding to the pattern that exhibits the highest similarity to the current system state. Interestingly, the boundaries separating these regions consistently intersect at the system's fixed point (indicated by the black circle). At the fixed point, all patterns receive nearly equal weights, indicating that the equilibrium is represented collectively rather than by a single representative pattern, e.g., see Fig.~\ref{fig:lotka-known-rhs-pattern-weight} for $P=4$. Although we do not yet have a theoretical explanation for this phenomenon, it is robust across all tested values of $P$.

Second, not every identified pattern dominates a distinct region of the state space. For example, when $P=16$, only six patterns are dominant in different region, while the remaining ten never attain the maximum similarity. Nevertheless, these seemingly inactive patterns remain important for prediction accuracy because the proposed framework combines information from all patterns through a weighted average rather than relying solely on the dominant one.

\subsubsection{Learning with unknown system dynamics}
Next, we consider the setting in which the governing equations are unavailable. This situation arises when only observations of the system are accessible or when the underlying solution operator is embedded within a complex or proprietary simulation code. In this case, we introduce a surrogate model (or emulator) to approximate either $f(\cdot)$ or $\mathcal{M}(\cdot)$ during the pattern learning stage. The surrogate may be trained independently prior to pattern learning or optimized jointly with the pattern representations. However, once the patterns have been identified, only the responses associated with the learned patterns are retained. Consequently, the surrogate model is no longer required during inference, and predictions can be performed using the stored pattern responses alone. 

\begin{figure}[ht!]
    \centering
    \includegraphics[width=0.8\linewidth]{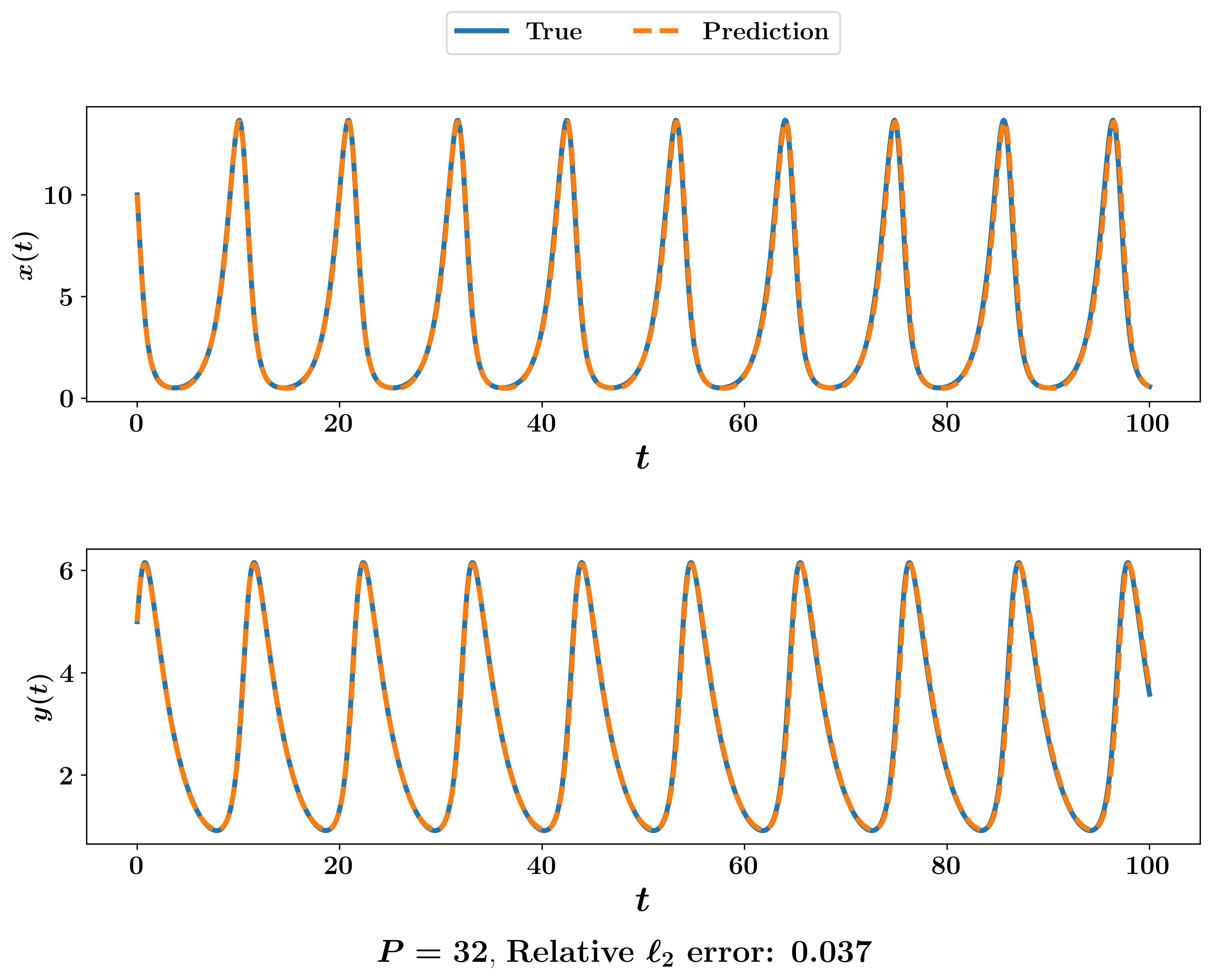}
    \caption{Prediction of the Lotka--Volterra system without access to the governing equations. A surrogate model is used to learn the flow map during pattern identification, after which the responses of the $32$ identified patterns are stored and used for recursive trajectory prediction.}
    \label{fig:lotka-unknown-rhs-learn-flowmap}
\end{figure}

Figure~\ref{fig:lotka-unknown-rhs-learn-flowmap} presents the prediction of the Lotka-Volterra system in the absence of the governing equations using $32$ identified patterns. We use a feedforward neural network with $3$ hidden layers, each layer has a width of $128$, to learn the flow map. This flow map is learned jointly with the patterns during the offline phase and we apply in-the-loop training with incremental increases in the roll-out length. Training both the flow map and the patterns takes about $3$ hours for a total of $500\text{K}$ optimization iterations. After training, only the patterns and the flow map responses associated with the identified patterns are retained for inference. Potentially, we can also restructure the training phase to directly learn the patterns and their local responses (instead of learning the global flow map), reducing the trainable parameter to $2nP$. Figure~\ref{fig:lotka-unknown-rhs-learn-flowmap} shows that the proposed framework accurately predicts both state variables over the entire prediction horizon, $t\in[0,100]$. In particular, the predicted trajectories preserve both the oscillation amplitudes and phase relationships of the reference solution, demonstrating that the learned local flow map representation faithfully captures the underlying nonlinear dynamics. The resulting relative $\ell_2$ error over the complete trajectory is around $3\%$, indicating that the recursive application of the learned pattern responses remains stable and accurate over long time horizons.

\subsection{Lorenz System}
The Lorenz system is a canonical nonlinear dynamical model that exhibits chaotic behavior and has been widely used as a benchmark for evaluating data-driven modeling and system identification approaches. Originally introduced as a simplified model of atmospheric convection, the system captures the emergence of complex dynamics from a set of low-dimensional nonlinear interactions. The governing equations are
\begin{equation}
\begin{aligned}
\frac{dx}{dt} &= \sigma(y-x),\\
\frac{dy}{dt} &= x(\rho-z)-y,\\
\frac{dz}{dt} &= xy-\beta z,
\end{aligned}
\end{equation}
where $x(t)$, $y(t)$, and $z(t)$ represent the system states, and $\sigma$, $\rho$, and $\beta$ are model parameters controlling the strength of coupling, thermal forcing, and dissipation, respectively. In this study, we use the standard chaotic parameter regime with $\sigma=10$, $\rho=28$, and $\beta=8/3$, which produces the characteristic butterfly-shaped attractor and sensitive dependence on initial conditions. Trajectories are generated through numerical integration from multiple initial conditions and are used to evaluate the performance of the proposed framework. Initial conditions are randomly sampled from a uniform distribution over the unit cube,
\begin{equation}
\mathbf{x}_0 \sim \mathcal{U}([0,1]^3), \label{eq:lorenz-ic}
\end{equation}
where $\mathbf{x}_0=(x_0,y_0,z_0)$ denotes the initial state of the Lorenz system.

Here, we consider the setting in which the governing equations are unavailable and only data samples of the system are accessible. Specifically, we employ a feedforward neural network as a surrogate model to learn the flow map and evaluate the output at the identified patterns. For training, we generate $10$ trajectories from distinct initial conditions sampled according to Eq.~\ref{eq:lorenz-ic}, with $t\in[0,10]$ and a time step of $0.01$. During testing, we generate an additional $10$ trajectories using independently sampled initial conditions from the same distribution to evaluate the predictive performance of the proposed framework.

\begin{figure}[ht!]
    \centering
    \includegraphics[width=0.8\linewidth]{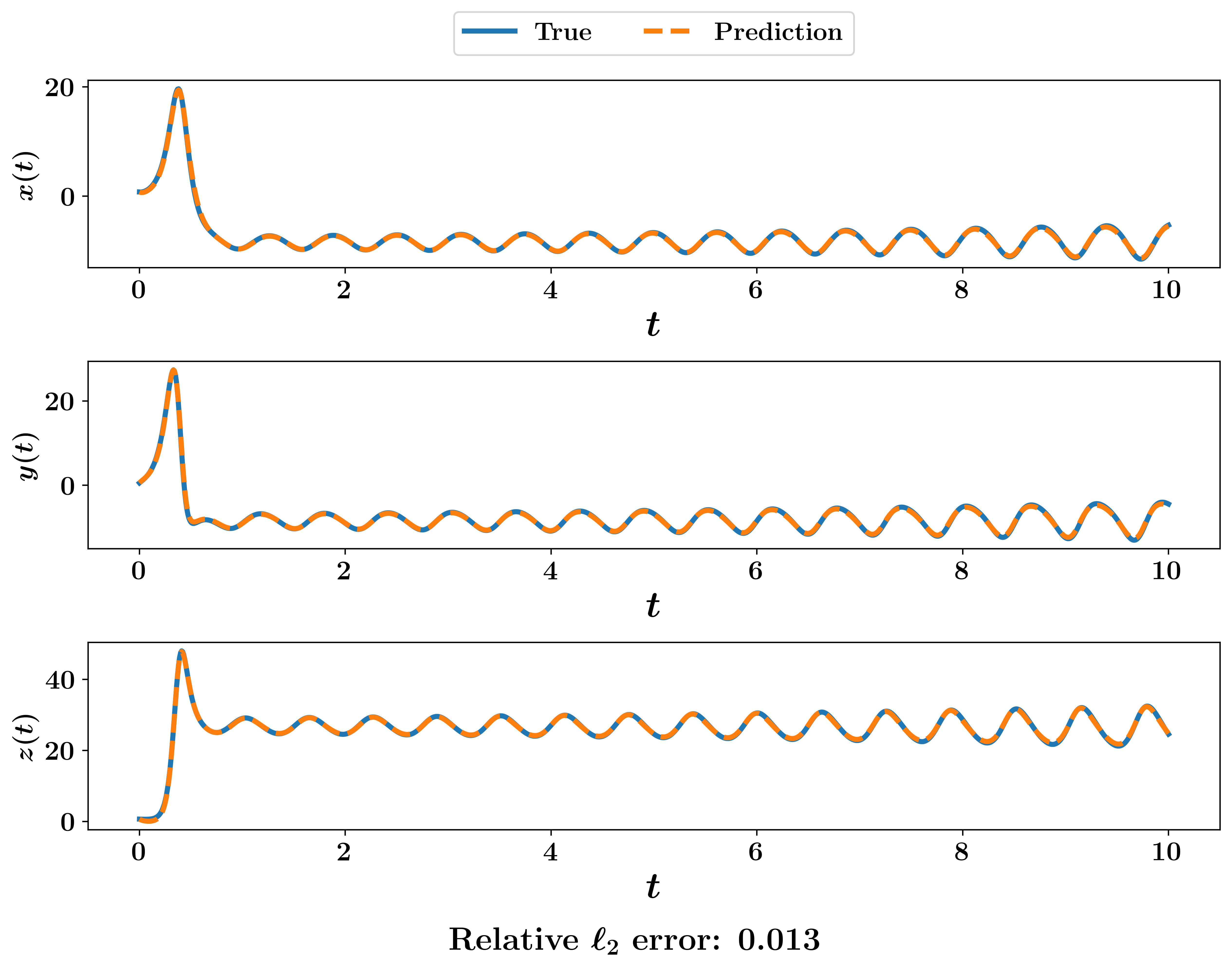}
    \caption{Long-term prediction of the Lorenz system trajectory from a test initial condition using $16$ identified patterns. The trajectory is generated recursively by repeatedly applying the learned flow map, starting from the initial condition at $t=0$.}
\label{fig:lorenz-baseline-prediction}
\end{figure}

\begin{figure}[ht!]
    \centering
    \includegraphics[width=\linewidth]{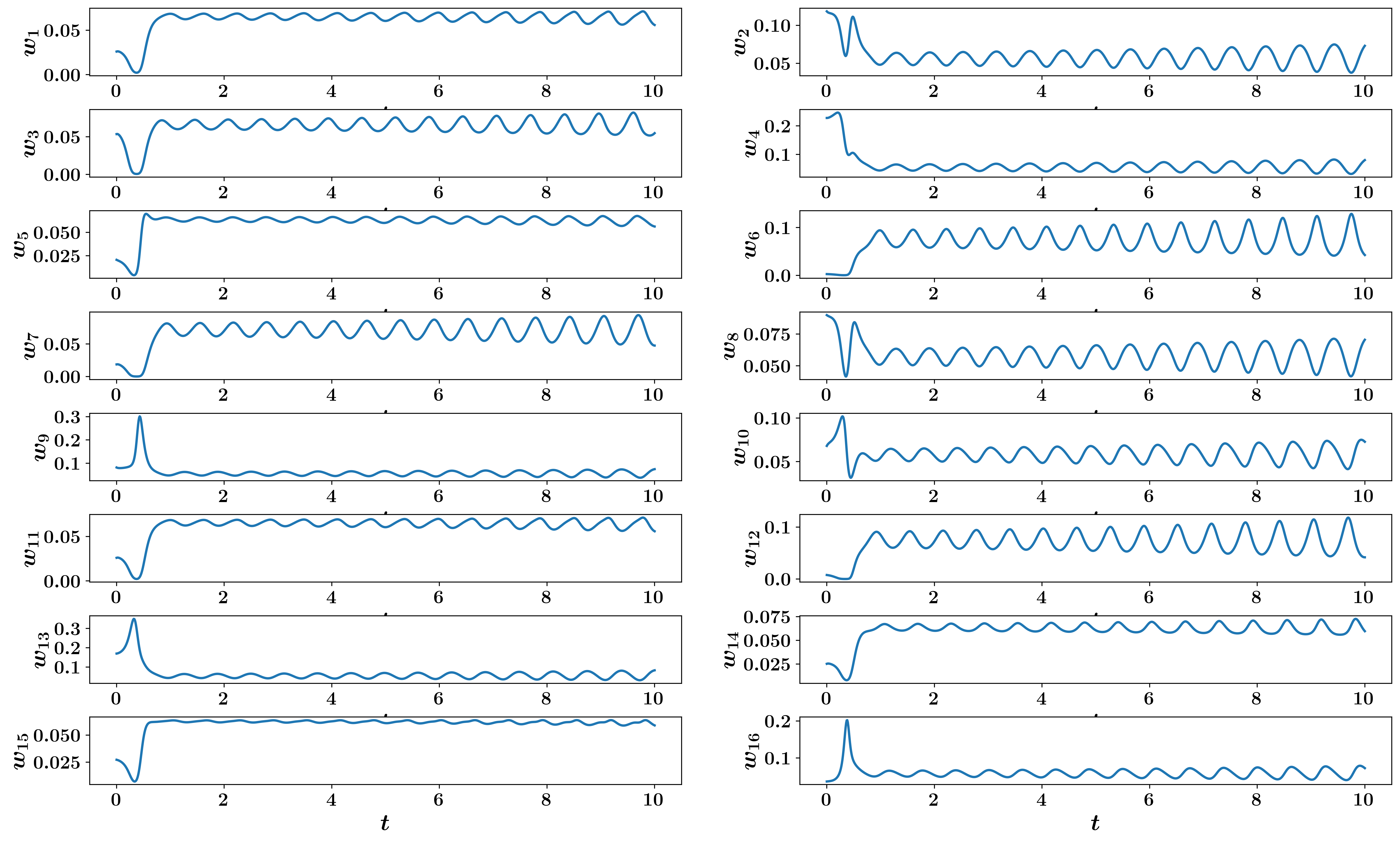}
    \caption{Temporal evolution of pattern weights during a Lorenz system trajectory. Different patterns become active at different stages of the dynamics.}
    \label{fig:lorenz-baseline-weight}
\end{figure}

Figure~\ref{fig:lorenz-baseline-prediction} presents the prediction results for one representative test trajectory using $16$ identified patterns. The average $\ell_2$ error across the test trajectories is approximately $1\%$. We emphasize that these predictions are obtained through recursive application of the learned flow map, starting from the initial condition at $t=0$. Figure~\ref{fig:lorenz-baseline-weight} illustrates the evolution of the similarity scores and corresponding pattern weights over time. The results show that different patterns become activated during distinct phases of the dynamics: a subset of patterns contributes more strongly during the initial transient transition, while other patterns become more active during the subsequent oscillatory motion.

\subsubsection{A continual learning perspective}
A common challenge in modeling nonlinear dynamical systems arises when the system evolves into dynamical regimes that were not represented during the initial training and pattern identification phase. Continual learning provides a natural mechanism for incorporating newly observed data to expand the model's representational capacity \cite{wang2024comprehensive,howard2024multifidelity}. However, many continual learning approaches suffer from catastrophic forgetting, whereby adapting to new data degrades performance on previously learned behaviors. To investigate this capability, we partition the Lorenz dataset into two sequential phases. In the first phase, training data are available over the interval $t\in[0,8]$, while the second phase augments the dataset with trajectory over $t\in[8,15]$. 

\begin{figure}[ht!]
    \centering
    \includegraphics[width=\linewidth]{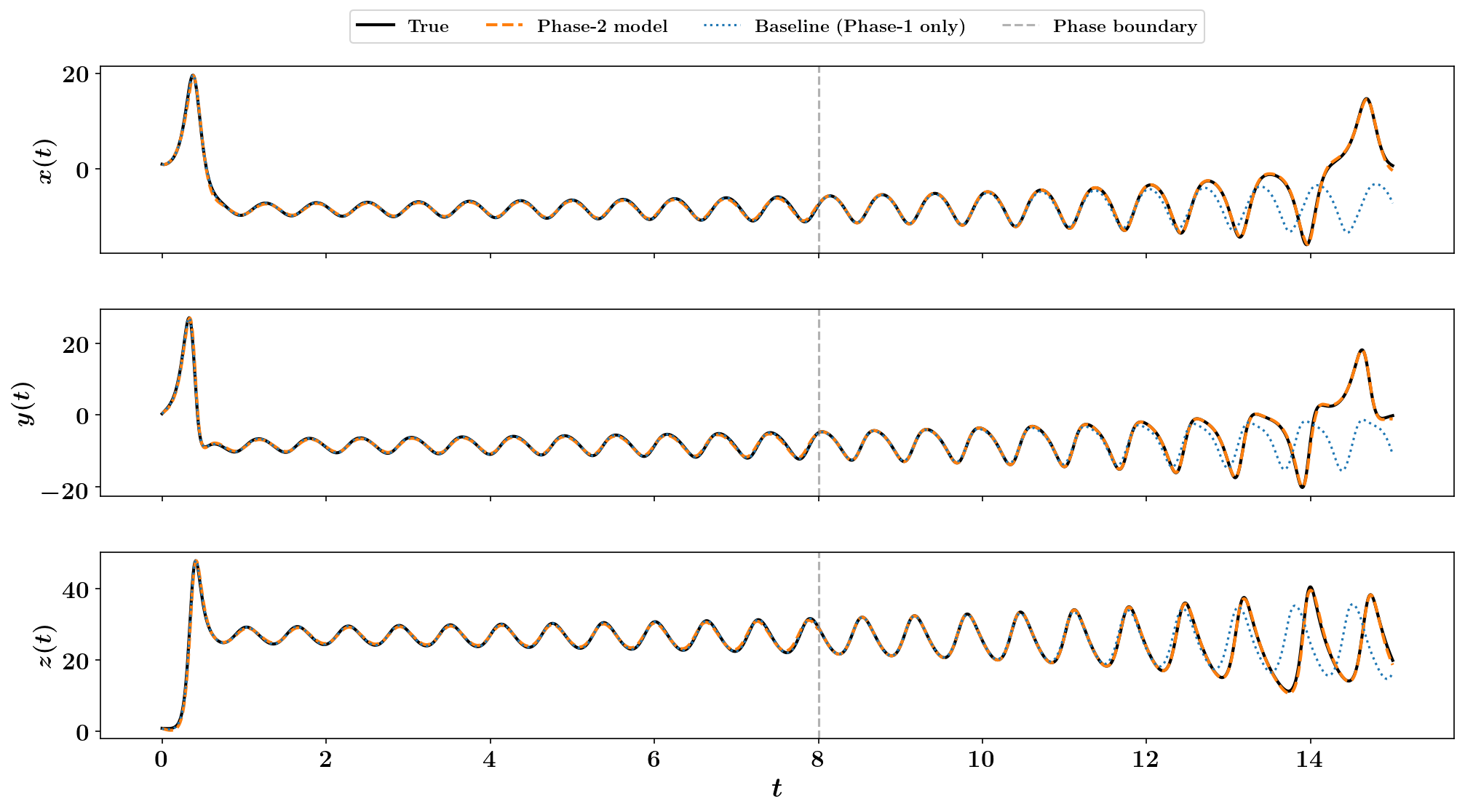}
    \caption{Continual learning through pattern augmentation. After training on data from $t\in[0,8]$, the original patterns are frozen and augmented with additional trainable patterns using data from $t\in[8,15]$. The augmented model successfully captures the subsequent transition that is missed by the baseline model.}
    \label{fig:lorenz-continual}
\end{figure}

The model is first trained using $16$ patterns to approximate the flow map over the first time interval. During the second phase, the learned patterns and their associated responses are frozen, and an additional $16$ trainable patterns are introduced to capture the newly observed dynamics. Figure~\ref{fig:lorenz-continual} compares trajectory predictions over the full time interval, $t\in[0,15]$, obtained using the baseline model trained only on the first phase and the continual learning model after incorporating the second phase. The baseline model fails to capture the transition near $t\approx14$, resulting in a relative $\ell_2$ error of $23.9\%$. In contrast, the continual learning model accurately predicts the transition, reducing the relative $\ell_2$ error to $1.8\%$. To isolate the effect of the learned pattern representation from the exponential error growth inherent to chaotic dynamics, both models are reinitialized with the true system state at $t=8$ before predicting the second phase.

\section{Applications to Regression} \label{sec:res-regression}
We present results for  regression using the approach outlined in Section \ref{sec:function_regression}. In Section \ref{sec:res-function regression} we present results for data-driven function regression while in Section \ref{sec:res-physics regression} we present results for physics-informed function regression.

\subsection{Data-driven function regression}\label{sec:res-function regression}

We begin with 3 examples for data-driven function regression. 

{\bf Example 1}. We consider the parabola given by $u(x)= 1- (2x-1)^2$ in $D=[0,1].$ We use $P=10$ patterns (classes), 50 training points and 50 test points. The training and test points are chosen uniformly in $D.$ The spatial positions for the 10 patterns are equidistant in $[0,1].$ The vectors used to compute the similarity of a query point to the pattern points have dimension $n=10.$ The variances for the influence functions are $\sigma_P^2=|D|/P=1/10$  and $\sigma_Q^2=|D|/n=1/10,$ respectively. The trainable values of the 10 patterns are found solving a least-squares problem and the test relative error is 0.1\%. The least-squares problems takes 0.01s to solve on a Macbook.

{\bf Example 2}. We consider a random combination of 10 sines with wavenumbers from 1 to 10 given by $u(x)= \sum_{i=1}^{10} a_i \sin(i2\pi x)$ in $D=[0,1],$ where $a_i = U[-1,1].$ We use $P=30$ patterns (classes), 250 training points and 250 test points. The training and test points are chosen uniformly in $D.$ The spatial positions for the 30 patterns are equidistant in $[0,1].$ The vectors used to compute the similarity of a query point to the pattern points have dimension $n=30.$ The variances for the influence functions are $\sigma_P^2=|D|/P=1/30$  and $\sigma_Q^2=|D|/n=1/30,$ respectively. The trainable values of the 30 patterns are found solving a least-squares problem and the test relative error is around 1\%. The least-squares problems takes 0.01s to solve on a Macbook.

{\bf Example 3}.We consider the function given by 
\begin{equation}
u(x)=
\begin{cases}
  1-(2(x-1)-1)^2 \text{ if } x \in [1,2], \\
  0 \text{ elsewhere.}  
\end{cases}
\end{equation}
We restrict the domain to $D=[0,3].$
We use $P=30$ patterns (classes), 250 training points and 250 test points. The training and test points are chosen uniformly in $D.$ The spatial positions for the 30 patterns are equidistant in $[0,3].$ The vectors used to compute the similarity of a query point to the pattern points have dimension $n=30.$ The variances for the influence functions are $\sigma_P^2=|D|/P=3/30$  and $\sigma_Q^2=|D|/n=3/30,$ respectively. The trainable values of the 30 patterns are found solving a least-squares problem and the test relative error is around 1\%. The least-squares problems takes 0.01s to solve on a Macbook.

\begin{figure}[ht!]
    \centering
\begin{subfigure}
[b]{.3\textwidth} 
\centering
\includegraphics[width=\linewidth]{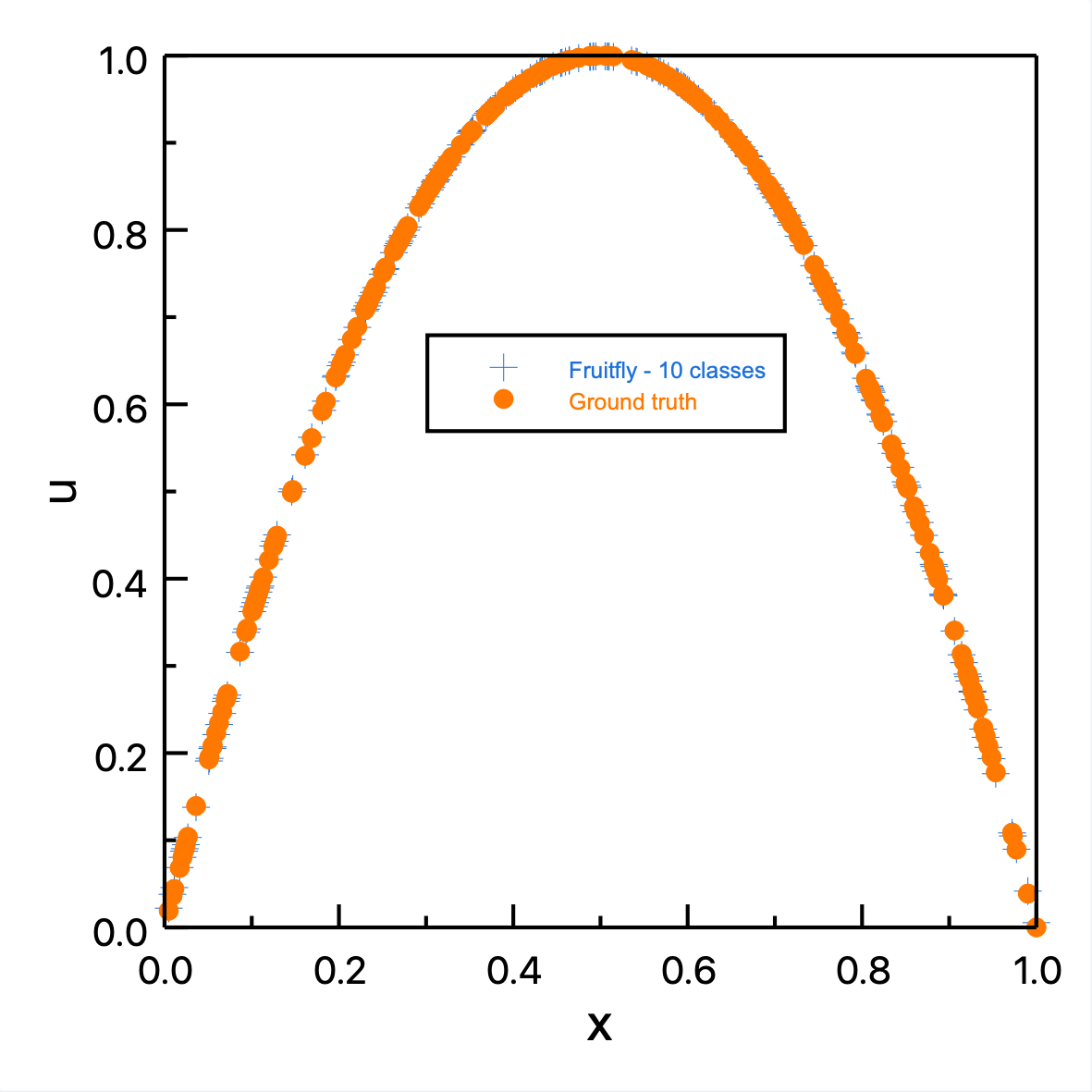}
  \end{subfigure}
\begin{subfigure}
[b]{.3\textwidth} 
\centering
\includegraphics[width=\linewidth]{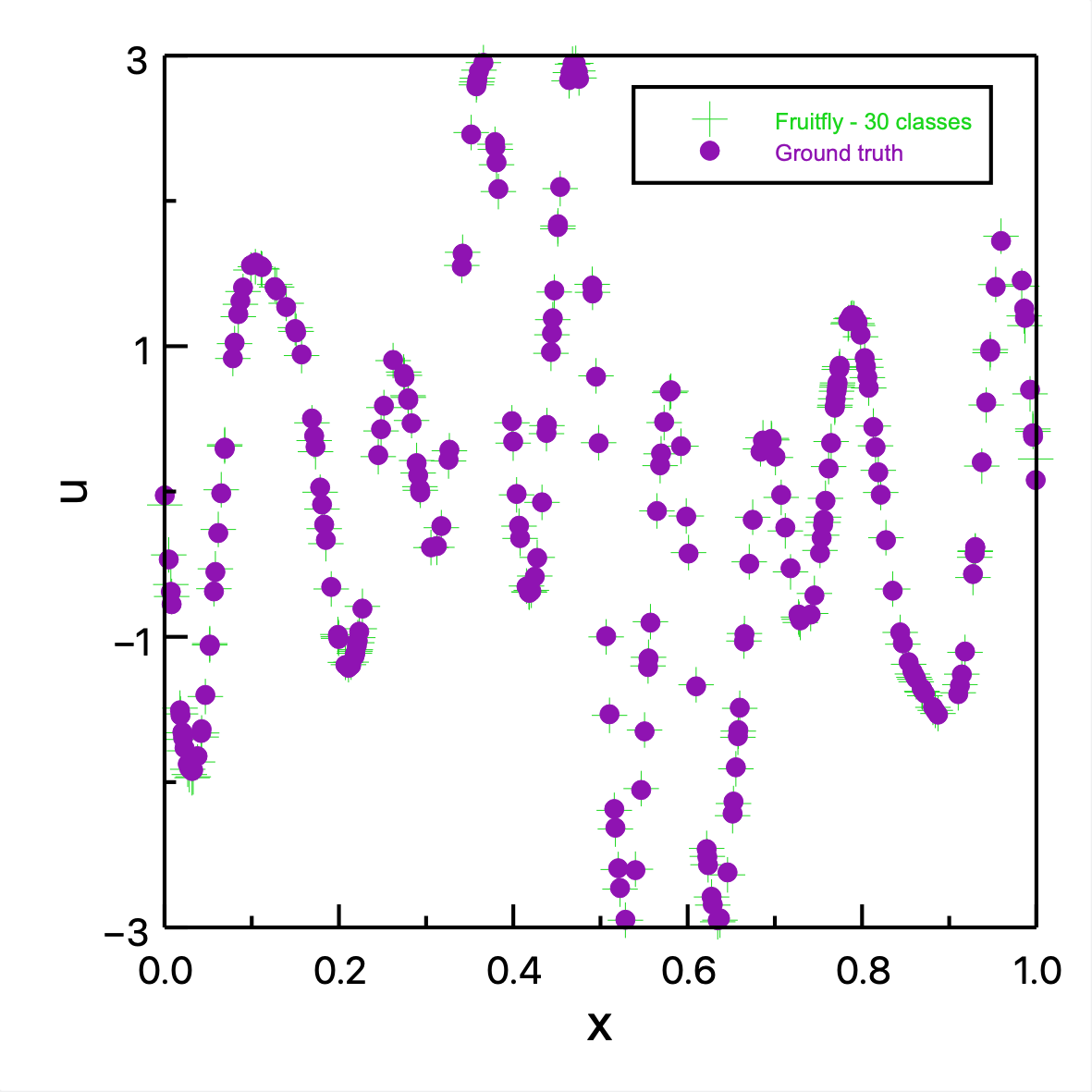}
  \end{subfigure} 
\begin{subfigure}[b]{.3\textwidth}
  \centering\includegraphics[width=\linewidth]{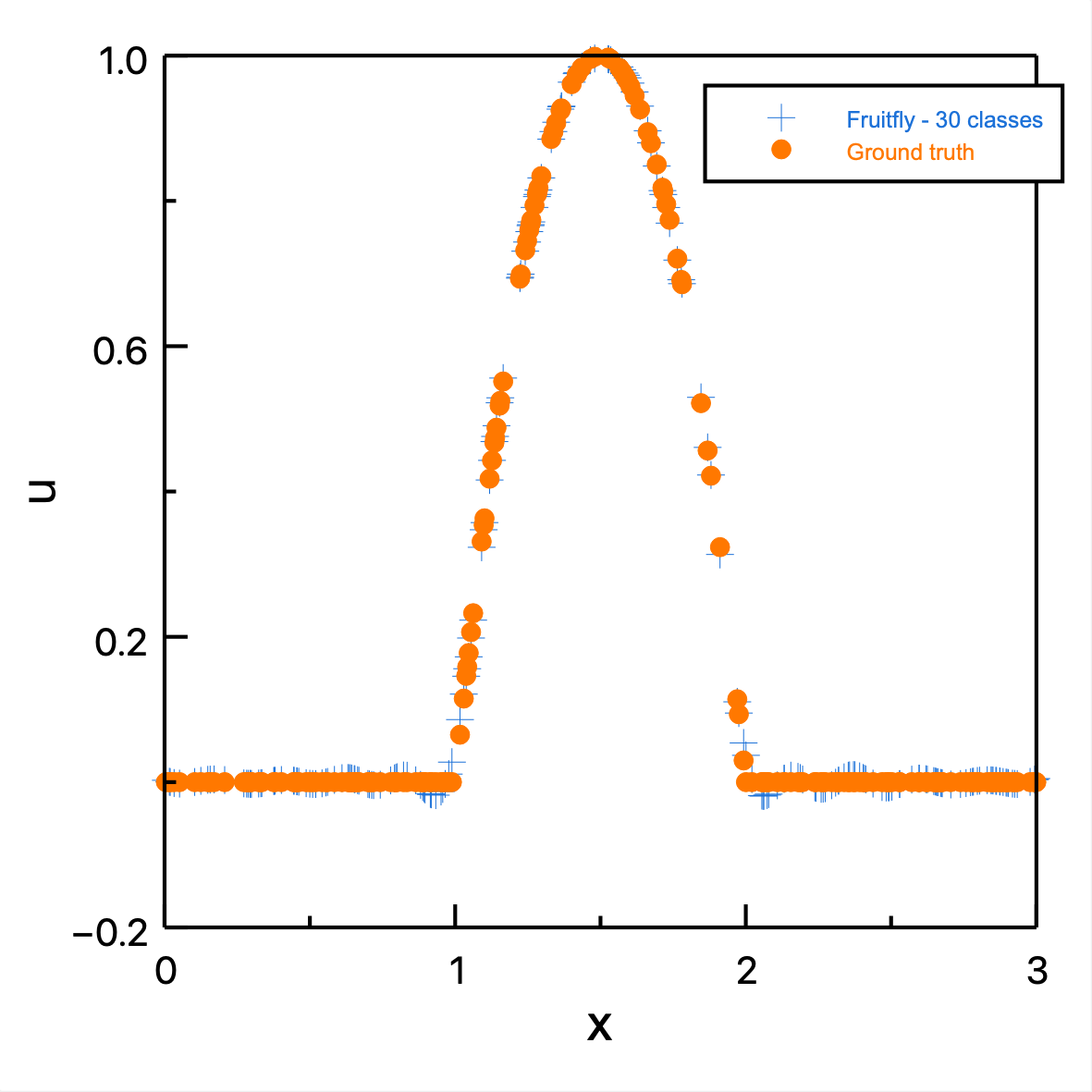}
  \end{subfigure}
    \caption{. Results for  function regression. (Left) Parabola. (Middle) A random combination of 10 sines with  wavenumbers from 1 to 10. (Right) Parabola with zero values outside of $[1,2].$}
    \label{fig:function_regression}
\end{figure}

Figure~\ref{fig:function_regression} shows the results of the fruitfly function regressor for the 3 examples.  

\subsection{Physics-informed regression}\label{sec:res-physics regression}

As proof-of-concept, we present results for the 1D Poisson equation  
\begin{equation}\label{poisson}
\frac{d^2u}{dx^2}=f(x),
\end{equation}
with zero Dirichlet conditions at the left and right boundary of the domain $D$ and a few choices of forcing functions $f(x).$ 

The loss function for the Poisson equation in the physics-informed case is given by

\begin{equation}\label{poisson_loss}
\mathcal{L}_{phys} = \frac{1}{N}\sum_{i=1}^N [\frac{d^2\hat{u}}{dx^2}(x_i)-f(x_i)]^2=\frac{1}{N}\sum_{i=1}^N [\sum_{k=1}^P \frac{d^2w_k}{dx^2}(x_i) \tilde{u}_{Pk} -f(x_i)]^2,
\end{equation}
where we have used the regressor estimate provided by \eqref{regressor_prediction}. Eq. \eqref{poisson_loss} can be used to estimate the  pattern values $\tilde{u}_{Pk},$ for $k=1,\ldots,P,$ by minimizing the value of $\mathcal{L}_{phys}.$  For the Poisson equation, this is a convex optimization problem which we can solve as a least-squares problem or with other optimizers. We opt here to present results when solving it as a least-squares problem. We also tried other approaches, like gradient descent and the Gauss-Newton method, and the obtained results are similar. 

{\bf Example 1}. We consider the forcing function $f(x)= 1- (2x-1)^2$ in $D=[0,1].$ We use $P=20$ patterns (classes), 50 training points and 50 test points. The training and test points are chosen uniformly in $D.$ The spatial positions for the 20 patterns are equidistant in $[0,1].$ The vectors used to compute the similarity of a query point to the pattern points have dimension $n=20.$ The variances for the influence functions are $\sigma_P^2=|D|/P=1/20$  and $\sigma_Q^2=|D|/n=1/20,$ respectively. The trainable values of the 20 patterns are found solving a least-squares problem and the test relative error is 1\%. The least-squares problems takes 0.01s to solve on a Macbook.

{\bf Example 2}. We consider the forcing function  $f(x)= \sum_{i=1}^{10} a_i \sin(i2\pi x)$ in $D=[0,1],$ where $a_i = U[-1,1].$ Note that the forcing function used here is a different combination than the one used in Example 2 of function regression. This is due to the different samples for the combination coefficients $a_i.$ We use $P=40$ patterns (classes), 250 training points and 250 test points. The training and test points are chosen uniformly in $D.$ The spatial positions for the 40 patterns are equidistant in $[0,1].$ The vectors used to compute the similarity of a query point to the pattern points have dimension $n=40.$ The variances for the influence functions are $\sigma_P^2=|D|/P=1/40$  and $\sigma_Q^2=|D|/n=1/40,$ respectively. The trainable values of the 40 patterns are found solving a least-squares problem and the test relative error is around 1\%. The least-squares problems takes 0.2s to solve on a Macbook.

{\bf Example 3}.We consider the forcing function  
\begin{equation}
f(x)=
\begin{cases}
  1-(2(x-1)-1)^2 \text{ if } x \in [1,2], \\
  0 \text{ elsewhere.}  
\end{cases}
\end{equation}
We restrict the domain to $D=[0,3].$
We use $P=100$ patterns (classes), 250 training points and 250 test points. The training and test points are chosen uniformly in $D.$ The spatial positions for the 100 patterns are equidistant in $[0,3].$ The vectors used to compute the similarity of a query point to the pattern points have dimension $n=100.$ The variances for the influence functions are $\sigma_P^2=|D|/P=3/100$  and $\sigma_Q^2=|D|/n=3/100,$ respectively. The trainable values of the 100 patterns are found solving a least-squares problem and the test relative error is around 10\%. Compared to the other cases, the increased relative error is understandable since the forcing function is not differentiable at $x=1$ and $x=2.$ Also, we have used equidistant positions where we evaluate the patterns, while a higher concentration of points close to $x=1,2$ would benefit the fruitfly regressor. The least-squares problems takes 3.5s to solve on a Macbook.

\begin{figure}[ht!]
    \centering
\begin{subfigure}
[b]{.3\textwidth} 
\centering
\includegraphics[width=\linewidth]{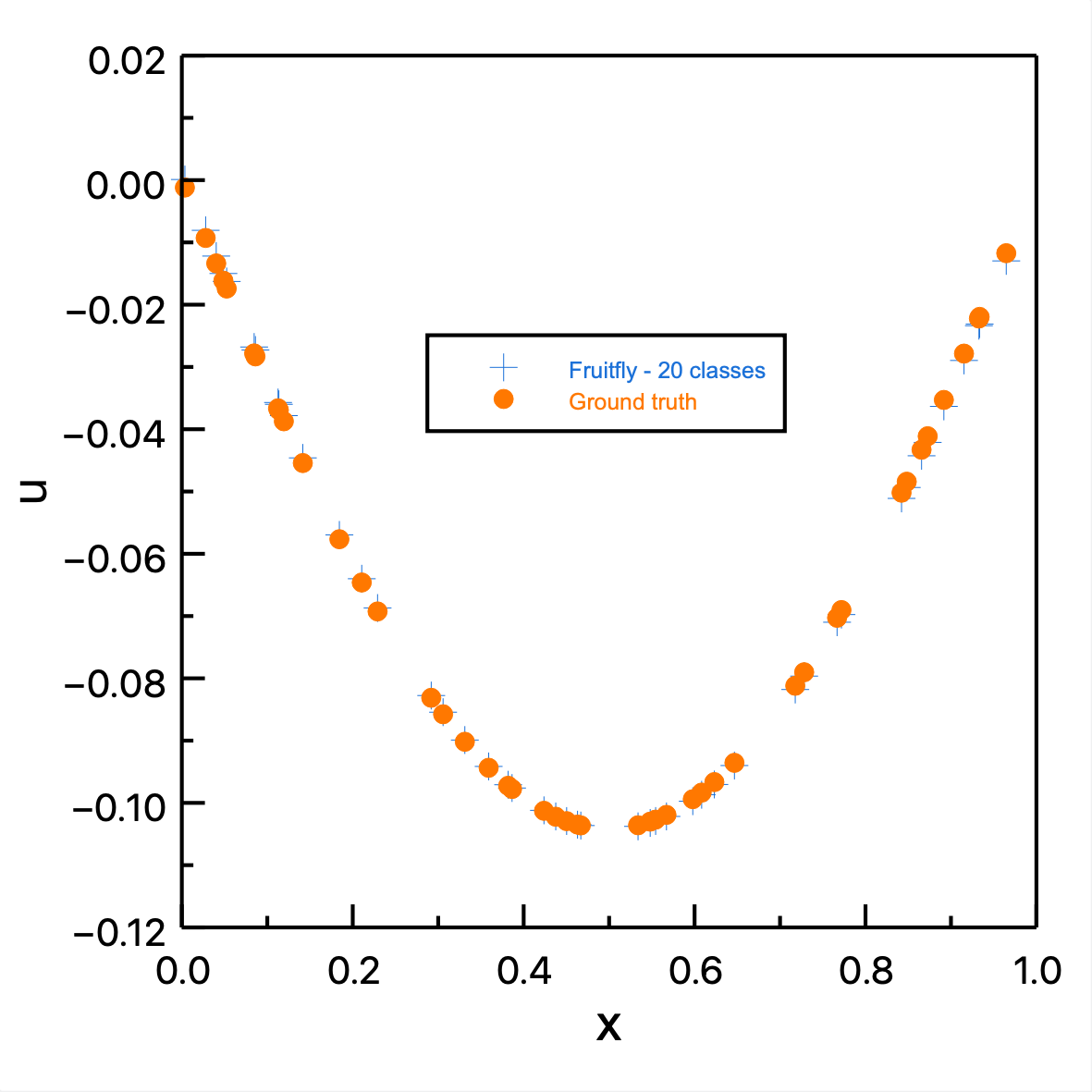}
  \end{subfigure}
\begin{subfigure}
[b]{.3\textwidth} 
\centering
\includegraphics[width=\linewidth]{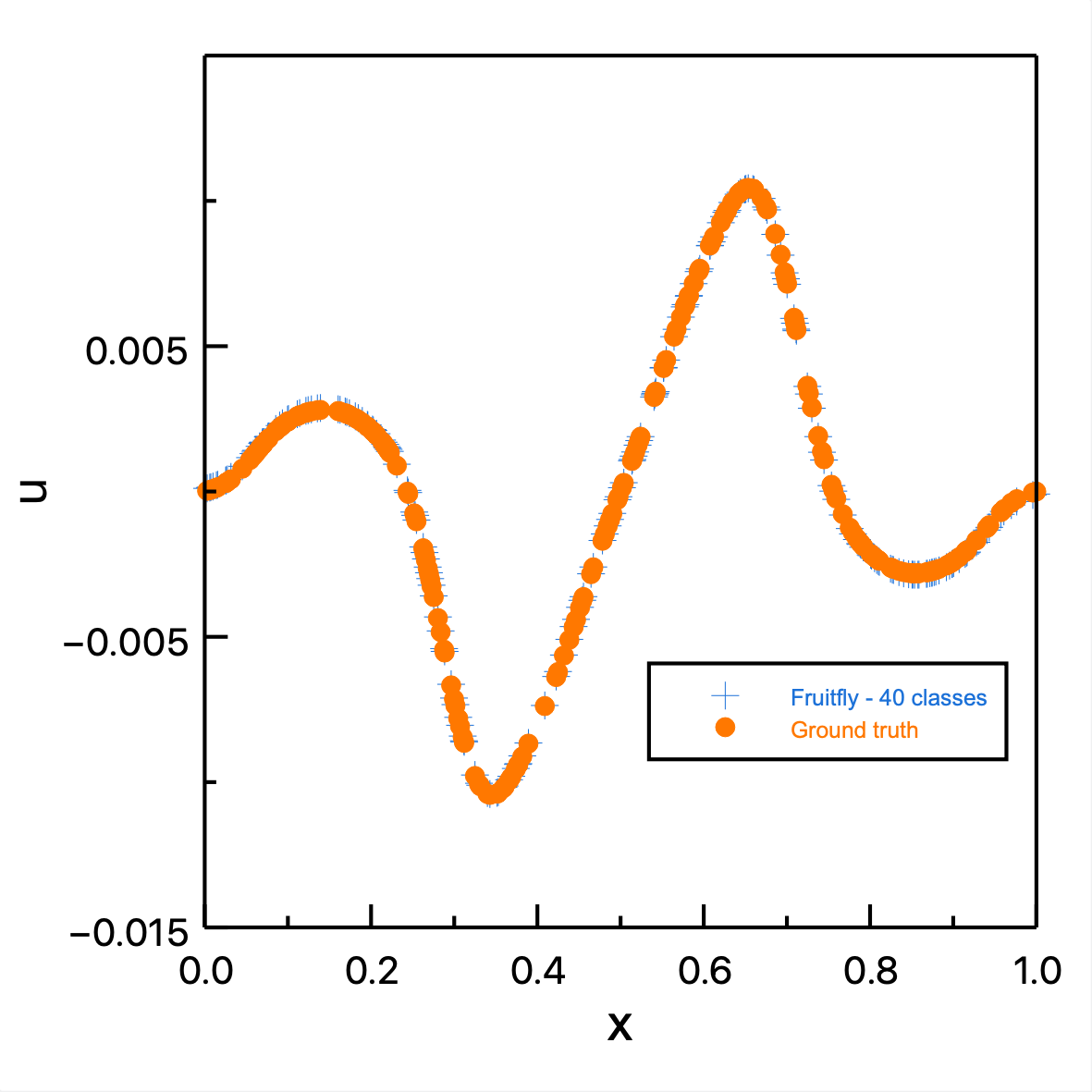}
  \end{subfigure} 
\begin{subfigure}[b]{.3\textwidth}
  \centering\includegraphics[width=\linewidth]{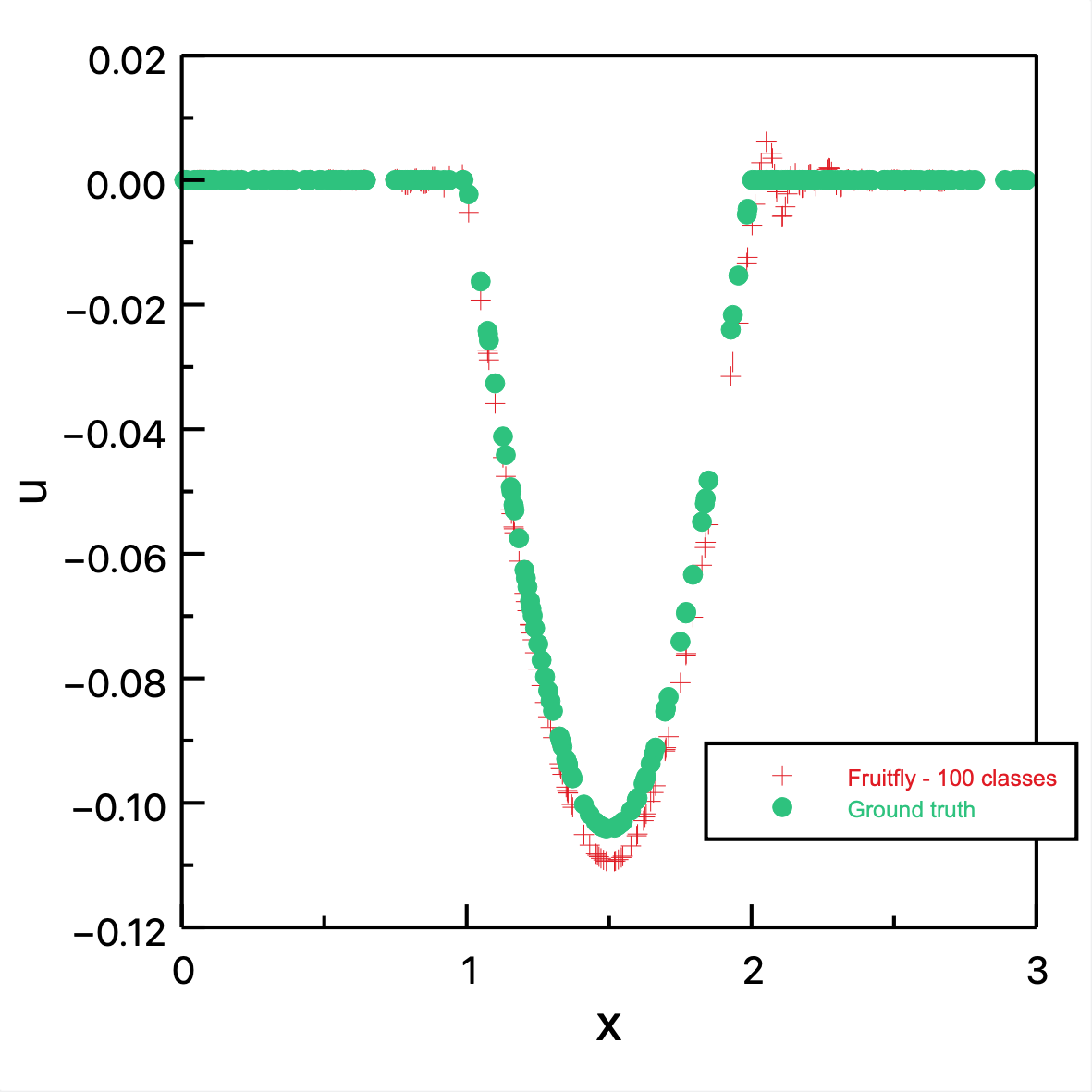}
  \end{subfigure}
    \caption{. Results for  1D Poisson with different forcing functions. (Left) Parabola. (Middle) A random combination of 10 sines with  wavenumbers from 1 to 10. (Right) Parabola with zero values outside of $[1,2].$}
    \label{fig:poisson_regression}
\end{figure}

Figure~\ref{fig:poisson_regression} shows the results of the fruitfly Poisson regressor for the 3 examples. 

\section{Discussion} \label{sec:conc}
Across all settings considered in this work, the proposed methodology follows the same computational pipeline. First, the input is represented either by the system state or by an application-specific embedding. Second, the input is softly classified among a finite collection of representative patterns. Finally, the desired output is reconstructed through a weighted combination of their associated stored responses. The quantity being reconstructed depends on the application. For continuous-time dynamics, it represents the governing right-hand-side operator. For discrete-time systems, it represents the one-step flow map. For general regression, it corresponds to the target nonlinear function. In every case, however, prediction is performed through the same classification-based reconstruction mechanism, providing a unified framework for approximating
nonlinear mappings across a broad class of problems.

In addition to application to more complex problems, there are different aspects of the proposed approach that warrant further investigation. First, the choice of number of patterns can be made adaptive with a criterion to decide when it is time to allow for more patterns. Second, the similarity, and possible redundancy of the learned patterns should be quantified and used to avoid the needless proliferation of patterns. Third, the locations of the pattern points can also be made part of the training process, allowing to focus more points in regions of the domain where the function to be regressed varies more rapidly. Fourth, for the function regression applications, either data-driven or physics-informed, the user prescribes the variance of the influence functions for the query and pattern points. In the numerical results we have included here, we have chosen the variance of the pattern and query points to be the same. Moreover, the variance was set equal to the resolution of the domain afforded by the pattern points i.e., $\frac{|D|}{P}.$ This choice resembles the variance of a random walker starting from the pattern and query points. It appears counterintuitive because one would expect that a variance equal to $(\frac{|D|}{P})^2,$ corresponding to ballistic movement, would be more appropriate. However, numerically we found that setting the variance equal to $\frac{|D|}{P}$ led to much more accurate results. A more thorough study of the choice of variance will be presented elsewhere.

We should also note that while the approach presented here advances classification as a way to aid regression, one can proceed in the opposite direction i.e., use regression to aid classification. In recent work by one of the current authors, a novel approach to classification was proposed which lifts the vectors to be classified to a higher dimensional space and then evolves them using a chaotic dynamical system before the final classification step \cite{stinis2026enhancing}. The approach provided significant improvement in terms of accuracy and efficiency compared to the baseline classifier. Taken together with the approach in the current work, we hope that these approaches point to the existence of a fertile interface between classification and regression that will be explored more in future work.

\section{Acknowledgments}
This work is partially supported by the U.S. Department of Energy, Office of Science, Advanced Scientific Computing Research program under the "LEADS: LEarning-Accelerated Domain Science" project (Project No. 85462). The work of S. E. A. was also supported by the U.S. Department of Energy, Office of Science, Advanced Scientific Computing Research program through the Pacific Northwest National Laboratory Distinguished Computational Mathematics Fellowship (Project No. 71268). Pacific Northwest National Laboratory is a multi-program national laboratory operated for the U.S. Department of Energy by Battelle Memorial Institute under Contract No. DE-AC05-76RL01830

\bibliographystyle{unsrtnat}
\bibliography{references}

\begin{thebibliography}{42}
\providecommand{\natexlab}[1]{#1}
\providecommand{\url}[1]{\texttt{#1}}
\expandafter\ifx\csname urlstyle\endcsname\relax
  \providecommand{\doi}[1]{doi: #1}\else
  \providecommand{\doi}{doi: \begingroup \urlstyle{rm}\Url}\fi

\bibitem[Lucia et~al.(2004)Lucia, Beran, and Silva]{lucia2004reduced}
David~J Lucia, Philip~S Beran, and Walter~A Silva.
\newblock Reduced-order modeling: new approaches for computational physics.
\newblock \emph{Progress in Aerospace Sciences}, 40\penalty0 (1-2):\penalty0 51--117, 2004.

\bibitem[Benner et~al.(2017)Benner, Ohlberger, Cohen, and Willcox]{benner2017model}
Peter Benner, Mario Ohlberger, Albert Cohen, and Karen Willcox.
\newblock \emph{Model reduction and approximation: theory and algorithms}.
\newblock SIAM, 2017.

\bibitem[Rowley and Dawson(2017)]{rowley2017model}
Clarence~W Rowley and Scott~TM Dawson.
\newblock Model reduction for flow analysis and control.
\newblock \emph{Annual Review of Fluid Mechanics}, 49:\penalty0 387--417, 2017.

\bibitem[Hofmann et~al.(2008)Hofmann, Sch{\"o}lkopf, and Smola]{hofmann2008kernel}
Thomas Hofmann, Bernhard Sch{\"o}lkopf, and Alexander~J Smola.
\newblock Kernel methods in machine learning.
\newblock 2008.

\bibitem[Shawe-Taylor and Cristianini(2004)]{shawe2004kernel}
John Shawe-Taylor and Nello Cristianini.
\newblock \emph{Kernel methods for pattern analysis}.
\newblock Cambridge University Press, 2004.

\bibitem[Williams and Rasmussen(1995)]{williams1995gaussian}
Christopher Williams and Carl Rasmussen.
\newblock Gaussian processes for regression.
\newblock \emph{Advances in neural information processing systems}, 8, 1995.

\bibitem[Schulz et~al.(2018)Schulz, Speekenbrink, and Krause]{schulz2018tutorial}
Eric Schulz, Maarten Speekenbrink, and Andreas Krause.
\newblock A tutorial on gaussian process regression: Modelling, exploring, and exploiting functions.
\newblock \emph{Journal of Mathematical Psychology}, 85:\penalty0 1--16, 2018.

\bibitem[Wang(2023)]{wang2023intuitive}
Jie Wang.
\newblock An intuitive tutorial to gaussian process regression.
\newblock \emph{Computing in Science \& Engineering}, 25\penalty0 (4):\penalty0 4--11, 2023.

\bibitem[Buhmann(2000)]{buhmann2000radial}
Martin~Dietrich Buhmann.
\newblock Radial basis functions.
\newblock \emph{Acta Numerica}, 9:\penalty0 1--38, 2000.

\bibitem[Flyer and Wright(2009)]{flyer2009radial}
Natasha Flyer and Grady~B Wright.
\newblock A radial basis function method for the shallow water equations on a sphere.
\newblock \emph{Proceedings: Mathematical, Physical and Engineering Sciences}, pages 1949--1976, 2009.

\bibitem[Bishop and Bishop(2023)]{bishop2023deep}
Christopher~M Bishop and Hugh Bishop.
\newblock \emph{Deep learning: Foundations and concepts}.
\newblock Springer Nature, 2023.

\bibitem[Prince(2023)]{prince2023understanding}
Simon~JD Prince.
\newblock \emph{Understanding deep learning}.
\newblock MIT press, 2023.

\bibitem[Chollet and Chollet(2021)]{chollet2021deep}
Francois Chollet and Fran{\c{c}}ois Chollet.
\newblock \emph{Deep learning with {Python}}.
\newblock simon and schuster, 2021.

\bibitem[Fefferman et~al.(2013)Fefferman, Mitter, and Narayanan]{fefferman2013testing}
Charles Fefferman, Sanjoy Mitter, and Hariharan Narayanan.
\newblock Testing the manifold hypothesis.
\newblock \emph{arXiv preprint arXiv:1310.0425}, 2013.

\bibitem[Packard et~al.(1980)Packard, Crutchfield, Farmer, and Shaw]{packard1980geometry}
Norman~H Packard, James~P Crutchfield, J~Doyne Farmer, and Robert~S Shaw.
\newblock Geometry from a time series.
\newblock \emph{Physical Review Letters}, 45\penalty0 (9):\penalty0 712, 1980.

\bibitem[Cleveland and Devlin(1988)]{cleveland1988locally}
William~S Cleveland and Susan~J Devlin.
\newblock Locally weighted regression: an approach to regression analysis by local fitting.
\newblock \emph{Journal of the American Statistical Association}, 83\penalty0 (403):\penalty0 596--610, 1988.

\bibitem[Levin(1998)]{levin1998approximation}
David Levin.
\newblock The approximation power of moving least-squares.
\newblock \emph{Mathematics of Computation}, 67\penalty0 (224):\penalty0 1517--1531, 1998.

\bibitem[Mirzaei(2015)]{mirzaei2015analysis}
Davoud Mirzaei.
\newblock Analysis of moving least squares approximation revisited.
\newblock \emph{Journal of Computational and Applied Mathematics}, 282:\penalty0 237--250, 2015.

\bibitem[Melenk and Babu{\v{s}}ka(1996)]{melenk1996partition}
Jens~M Melenk and Ivo Babu{\v{s}}ka.
\newblock The partition of unity finite element method: basic theory and applications.
\newblock \emph{Computer Methods in Applied Mechanics and Engineering}, 139\penalty0 (1-4):\penalty0 289--314, 1996.

\bibitem[Babu{\v{s}}ka and Melenk(1997)]{babuvska1997partition}
Ivo Babu{\v{s}}ka and Jens~M Melenk.
\newblock The partition of unity method.
\newblock \emph{International Journal for Numerical Methods in Engineering}, 40\penalty0 (4):\penalty0 727--758, 1997.

\bibitem[Fan et~al.(2023)Fan, Trask, D'Elia, and Darve]{fan2023probabilistic}
Tiffany Fan, Nathaniel Trask, Marta D'Elia, and Eric Darve.
\newblock Probabilistic partition of unity networks for high-dimensional regression problems.
\newblock \emph{International Journal for Numerical Methods in Engineering}, 124\penalty0 (10):\penalty0 2215--2236, 2023.

\bibitem[Ferrari-Trecate et~al.(2003)Ferrari-Trecate, Muselli, Liberati, and Morari]{ferrari2003clustering}
Giancarlo Ferrari-Trecate, Marco Muselli, Diego Liberati, and Manfred Morari.
\newblock A clustering technique for the identification of piecewise affine systems.
\newblock \emph{Automatica}, 39\penalty0 (2):\penalty0 205--217, 2003.

\bibitem[Breschi et~al.(2016)Breschi, Piga, and Bemporad]{breschi2016piecewise}
Valentina Breschi, Dario Piga, and Alberto Bemporad.
\newblock Piecewise affine regression via recursive multiple least squares and multicategory discrimination.
\newblock \emph{Automatica}, 73:\penalty0 155--162, 2016.

\bibitem[Bemporad(2022)]{bemporad2022piecewise}
Alberto Bemporad.
\newblock A piecewise linear regression and classification algorithm with application to learning and model predictive control of hybrid systems.
\newblock \emph{IEEE Transactions on Automatic Control}, 68\penalty0 (6):\penalty0 3194--3209, 2022.

\bibitem[Amsallem et~al.(2012)Amsallem, Zahr, and Farhat]{amsallem2012nonlinear}
David Amsallem, Matthew~J Zahr, and Charbel Farhat.
\newblock Nonlinear model order reduction based on local reduced-order bases.
\newblock \emph{International Journal for Numerical Methods in Engineering}, 92\penalty0 (10):\penalty0 891--916, 2012.

\bibitem[Ahmed and San(2020)]{ahmed2020breaking}
Shady~E Ahmed and Omer San.
\newblock Breaking the kolmogorov barrier in model reduction of fluid flows.
\newblock \emph{Fluids}, 5\penalty0 (1):\penalty0 26, 2020.

\bibitem[Liang et~al.(2021)Liang, Ryali, Hoover, Grinberg, Navlakha, Zaki, and Krotov]{liang2021can}
Yuchen Liang, Chaitanya~K Ryali, Benjamin Hoover, Leopold Grinberg, Saket Navlakha, Mohammed~J Zaki, and Dmitry Krotov.
\newblock Can a fruit fly learn word embeddings?
\newblock \emph{arXiv preprint arXiv:2101.06887}, 2021.

\bibitem[Sauer et~al.(1991)Sauer, Yorke, and Casdagli]{sauer1991embedology}
Tim Sauer, James~A Yorke, and Martin Casdagli.
\newblock Embedology.
\newblock \emph{Journal of Statistical Physics}, 65\penalty0 (3):\penalty0 579--616, 1991.

\bibitem[Jacobs et~al.(1991)Jacobs, Jordan, Nowlan, and Hinton]{jacobs1991adaptive}
Robert~A Jacobs, Michael~I Jordan, Steven~J Nowlan, and Geoffrey~E Hinton.
\newblock Adaptive mixtures of local experts.
\newblock \emph{Neural Computation}, 3\penalty0 (1):\penalty0 79--87, 1991.

\bibitem[Shazeer et~al.(2017)Shazeer, Mirhoseini, Maziarz, Davis, Le, Hinton, and Dean]{shazeer2017outrageously}
Noam Shazeer, Azalia Mirhoseini, Krzysztof Maziarz, Andy Davis, Quoc Le, Geoffrey Hinton, and Jeff Dean.
\newblock Outrageously large neural networks: The sparsely-gated mixture-of-experts layer.
\newblock \emph{arXiv preprint arXiv:1701.06538}, 2017.

\bibitem[Sharma and Shankar(2024)]{sharma2024ensemble}
Ramansh Sharma and Varun Shankar.
\newblock Ensemble and mixture-of-experts {DeepONets} for operator learning.
\newblock \emph{arXiv preprint arXiv:2405.11907}, 2024.

\bibitem[Torgo and Gama(1996)]{torgo1996regression}
Lu{\'\i}s Torgo and Joao Gama.
\newblock Regression by classification.
\newblock In \emph{Brazilian symposium on artificial intelligence}, pages 51--60. Springer, 1996.

\bibitem[Salman and Kecman(2012)]{salman2012regression}
Raied Salman and Vojislav Kecman.
\newblock Regression as classification.
\newblock In \emph{2012 Proceedings of IEEE Southeastcon}, pages 1--6. IEEE, 2012.

\bibitem[Kaiser et~al.(2013)Kaiser, Noack, Cordier, Spohn, Segond, Abel, Daviller, {\"O}sth, Krajnovi{\'c}, and Niven]{kaiser2013cluster}
Eurika Kaiser, Bernd~R Noack, Laurent Cordier, Andreas Spohn, Marc Segond, Markus Abel, Guillaume Daviller, Jan {\"O}sth, Sini{\v{s}}a Krajnovi{\'c}, and Robert~K Niven.
\newblock Cluster-based reduced-order modelling of a mixing layer.
\newblock \emph{arXiv preprint arXiv:1309.0524}, 2013.

\bibitem[Fernex et~al.(2021)Fernex, Noack, and Semaan]{fernex2021cluster}
Daniel Fernex, Bernd~R Noack, and Richard Semaan.
\newblock Cluster-based network modeling—from snapshots to complex dynamical systems.
\newblock \emph{Science Advances}, 7\penalty0 (25):\penalty0 eabf5006, 2021.

\bibitem[Ahmed and Stinis(2023)]{ahmed2023multifidelity}
Shady~E Ahmed and Panos Stinis.
\newblock A multifidelity deep operator network approach to closure for multiscale systems.
\newblock \emph{Computer Methods in Applied Mechanics and Engineering}, 414:\penalty0 116161, 2023.

\bibitem[Raissi et~al.(2019)Raissi, Perdikaris, and Karniadakis]{raissi2019physics}
Maziar Raissi, Paris Perdikaris, and George~Em Karniadakis.
\newblock Physics-informed neural networks: A deep learning framework for solving forward and inverse problems involving nonlinear partial differential equations.
\newblock \emph{Journal of Computational Physics}, 378:\penalty0 686--707, 2019.
\newblock \doi{10.1016/j.jcp.2018.10.045}.

\bibitem[Cai(2001)]{cai2001weighted}
Zongwu Cai.
\newblock Weighted {Nadaraya--Watson} regression estimation.
\newblock \emph{Statistics \& Probability Letters}, 51\penalty0 (3):\penalty0 307--318, 2001.

\bibitem[Larsson et~al.(2017)Larsson, Shcherbakov, and Heryudono]{larsson2017least}
Elisabeth Larsson, Victor Shcherbakov, and Alfa Heryudono.
\newblock A least squares radial basis function partition of unity method for solving pdes.
\newblock \emph{SIAM Journal on Scientific Computing}, 39\penalty0 (6):\penalty0 A2538--A2563, 2017.

\bibitem[Wang et~al.(2024)Wang, Zhang, Su, and Zhu]{wang2024comprehensive}
Liyuan Wang, Xingxing Zhang, Hang Su, and Jun Zhu.
\newblock A comprehensive survey of continual learning: Theory, method and application.
\newblock \emph{IEEE Transactions on Pattern Analysis and Machine Intelligence}, 46\penalty0 (8):\penalty0 5362--5383, 2024.

\bibitem[Howard et~al.(2024)Howard, Fu, and Stinis]{howard2024multifidelity}
Amanda Howard, Yucheng Fu, and Panos Stinis.
\newblock A multifidelity approach to continual learning for physical systems.
\newblock \emph{Machine Learning: Science and Technology}, 5\penalty0 (2):\penalty0 025042, 2024.

\bibitem[Stinis(2026)]{stinis2026enhancing}
Panos Stinis.
\newblock Enhancing classification accuracy through chaos.
\newblock \emph{arXiv preprint arXiv:2603.15299}, 2026.

\end{thebibliography}

\end{document}